\documentclass{article}

\usepackage{microtype}
\usepackage{graphicx}
\usepackage{subfigure}
\usepackage{booktabs} 

\usepackage{amsmath,amsfonts,bm}
\usepackage{ulem}
\usepackage{wrapfig}
\usepackage{pifont} 

\usepackage{xspace}
\usepackage{ulem}
\usepackage[dvipsnames,svgnames]{xcolor}
\usepackage{wrapfig,lipsum,booktabs}
\usepackage{hyperref}
\usepackage{url}
\usepackage{graphicx}
\usepackage{multirow}
\usepackage{tikz}
\usepackage{color}
\usetikzlibrary{bayesnet}
\usepackage{graphicx}
\usepackage{color}
\usepackage{amsmath}
\usepackage{floatflt}
\usepackage{soul}
\usepackage{xspace}
\usepackage{subfigure}
\usepackage{capt-of}
\usepackage{algorithm}
\usepackage{algorithmic}
\usepackage[algo2e,ruled,vlined]{algorithm2e}

\usepackage[dynnworkshop]{icml2022}

\usepackage{amsmath}
\usepackage{amssymb}
\usepackage{mathtools}
\usepackage{amsthm}

\usepackage[capitalize,noabbrev]{cleveref}

\theoremstyle{plain}

\theoremstyle{definition}

\theoremstyle{remark}

\usepackage{paralist}

\usepackage[textsize=tiny]{todonotes}

\usepackage{array, booktabs, makecell}

\usepackage{booktabs}
\usepackage{array,tabularx}

\usepackage{pgfplots}
\pgfplotsset{compat=1.17}  
\usepackage{amsmath,amsfonts,bm}
\usepackage{xspace}
\usepackage{ulem}
\usepackage{wrapfig}

\usepackage{pifont} 
\newcommand{\bfilter}{\mathcal{B}}
\newcommand{\bfilterm}{$\mathcal{B}$\xspace}

\newcommand{\ours}{\textsc{flowgen}\xspace}

\newcommand{\eg}{e.g.,\xspace}
\newcommand{\ie}{i.e.,\xspace}

\newcommand{\gptz}{\textsc{gpt2}\xspace}

\newcommand{\coraml}{\textsc{coraml}\xspace}
\newcommand{\polblogs}{\textsc{polblogs}\xspace}

\newcommand{\pubmed}{\textsc{pubmed}\xspace}
\newcommand{\citeseer}{\textsc{citeseer}\xspace}
\newcommand{\auc}{\textsc{auc}\xspace}
\newcommand{\ap}{\textsc{ap}\xspace}

\newcommand{\squishlist}{
  \begin{list}{$\bullet$}
    { \setlength{\itemsep}{0pt}      \setlength{\parsep}{3pt}
      \setlength{\topsep}{3pt}       \setlength{\partopsep}{0pt}
      \setlength{\leftmargin}{1.5em} \setlength{\labelwidth}{1em}
      \setlength{\labelsep}{0.5em} } }
\newcommand{\reallysquishlist}{
  \begin{list}{$\bullet$}
    { \setlength{\itemsep}{0pt}    \setlength{\parsep}{0pt}
      \setlength{\topsep}{0pt}     \setlength{\partopsep}{0pt}
      \setlength{\leftmargin}{0.2em} \setlength{\labelwidth}{0.2em}
      \setlength{\labelsep}{0.2em} } }

 \newcommand{\squishend}{
     \end{list} 
 }

\newcommand{\aow}[1]{{\textcolor{black}{#1}}}
\newcommand{\eat}[1]{{}}

\newcommand{\yy}[1]{}

\newcommand{\TODO}[1]{}

\def\Secref#1{Section~\ref{#1}}

\def\eqref#1{equation~\ref{#1}}

\def\1{\bm{1}}

\def\vw{{\bm{w}}}

\def\mH{{\bm{H}}}

\def\mS{{\bm{S}}}

\def\mW{{\bm{W}}}

\DeclareMathAlphabet{\mathsfit}{\encodingdefault}{\sfdefault}{m}{sl}
\SetMathAlphabet{\mathsfit}{bold}{\encodingdefault}{\sfdefault}{bx}{n}

\def\gG{{\mathcal{G}}}

\def\sN{{\mathbb{N}}}

\def\sS{{\mathbb{S}}}

\def\sV{{\mathbb{V}}}

\newcommand{\pcond}[2]{p(#1 \mid #2)}

\DeclareMathOperator*{\argmax}{arg\,max}

\newcommand{\nhashf}{h}
\newcommand{\probfp}{P}
\newcommand{\nbits}{M}

\newcommand{\graph}{\gG}
\newcommand{\node}{v}
\newcommand{\nodeset}{\sV}

\newcommand{\maxdegree}{d_{\text{max}}}

\newcommand{\lentrainwalk}{k}
\newcommand{\lentestwalk}{l}
\newcommand{\ntrainwalks}{m}
\newcommand{\ntestwalks}{n}
\newcommand{\rwalk}{\vw}
\newcommand{\matwalk}{\mW}

\newcommand{\rwalkexp}{$\node_1, \node_2, \ldots, \node_{\lentrainwalk}$\xspace}

 \newcommand{\txt}[1]{$#1$\xspace}

\newcommand{\txtnode}{\txt{\node}}
\newcommand{\txtgraph}{\txt{\graph}}
\newcommand{\txtlentrainwalk}{\txt{\lentrainwalk}}
\newcommand{\txtlentestwalk}{\txt{\lentestwalk}}

\newcommand{\txtntrainwalks}{\txt{\ntrainwalks}}
\newcommand{\txtntestwalks}{\txt{\ntestwalks}}
\newcommand{\txtrwalk}{\txt{\rwalk}}
\newcommand{\txtmatwalk}{\txt{\matwalk}}

\newcommand{\fast}{\textsc{fast}\xspace}
\newcommand{\slow}{\textsc{slow}\xspace}
\newcommand{\fastslow}{\textsc{flow}\xspace}
\newcommand{\netgan}{NetGAN\xspace}

\newcommand{\handover}{$\node_j$\xspace}
\newcommand{\handoveridx}{j}

\newcommand{\nbrsize}{p}
\newcommand{\nbrsizetxt}{$p$\xspace}
\newcommand{\nbr}{\mathcal{N}}
\newcommand{\adj}{Adj}
\newcommand{\sa}{\textsc{sa}\xspace}
\icmltitlerunning{\ours: Fast and slow graph generation}

\begin{document}

\twocolumn[
\icmltitle{\ours: Fast and slow graph generation}



\icmlsetsymbol{equal}{*}

\begin{icmlauthorlist}
\icmlauthor{Aman Madaan}{comp}
\icmlauthor{Yiming Yang}{comp}
\end{icmlauthorlist}

\icmlaffiliation{comp}{Language Technologies Institute, Carnegie Mellon University, Pittsburgh, PA}

\icmlcorrespondingauthor{Aman Madaan}{amadaan@cs.cmu.edu}

\icmlkeywords{Machine Learning, ICML}

\vskip 0.3in
]



\printAffiliationsAndNotice{} 

\begin{abstract}
Machine learning systems typically apply the same model to both easy and tough cases. 
This is in stark contrast with humans, who tend to evoke either \textit{fast} (instinctive) or \textit{slow} (analytical) thinking process, depending on the difficulty of the problem---a property called the dual-process theory of mind.
We present \ours, a graph-generation model inspired by the dual-process theory of mind. 
Depending on the difficulty of graph completion at the current step, the system either calls a \fast~(weaker) module or a \slow~(stronger) module for the task. 
These modules have identical architectures, but vary in the number of parameters and consequently differ in generative power.
Experiments on real-world graphs show that \ours can successfully generate graphs similar to those generated by a single large model, while being up to 2x faster.
\end{abstract}

\section{Introduction}

Graphs provide a rich abstraction for a wide range of tasks including molecular design~\citep{de_cao_molgan_2018,samanta_nevae_2019,lim_scaffold-based_2020}, temporal and commonsense reasoning~\citep{madaan2020neural,madaan2021think,sakaguchi_proscript_2021,saha_explagraphs_2021}, online user interaction modeling~\citep{zhou_data-driven_2020}, and map layout design~\citep{mi_hdmapgen_2021}.
Developing generative models of graphs is is therefore an important classical problem, which has seen renewed interest with the success of deep learning models.
Specifically, \textit{implicit} generative models are a popular choice for graph generative modeling.
Unlike explicit models, implicit generative models do not explicitly model the distribution of graphs but instead allow sampling graphs.
A popular example of such implicit models are GANs, and have recently shown state of the art results for generative modeling of graphs~\citep{bojchevski2018netgan}.

Like typical machine learning models, generative models of graphs currently use identical model complexity and computational strength while generating graphs.
However, since these models are constructive by design~(\ie they build a graph piece-by-piece), it is natural to expect that generating different parts of a graph requires different levels of reasoning.
For example, generating a 2-hop neighborhood frequently seen during training might be \textit{easier} than generating a novel 4-hop neighborhood.


Indeed, it has long been posited~\citep{Posner1975AttentionAC, Shiffrin1977ControlledAA, evans_heuristic_1984, stanovich_individual_2000, kahneman_maps_2003,frankish_dual-process_2010} that humans frequently use differential reasoning based on the problem difficulty. 
For example, consider two problems: i) \textit{2 * 2 = ?}, and ii) \textit{203 * 197 = ?}
Both these problems involve multiplication between two integers. 
Yet, they pose a very different level of difficulty for a human solver.
The answer to 2*2 will almost instinctively come to most, while solving 19*3 will require more careful thinking.
Specifically, \citet{stanovich_individual_2000} propose to divide mental processing as being done by two metaphorical systems referred by them as \textit{System 1}~(instinctive, used for 2 * 2) and \textit{System 2}~(analytical, planner, used for 203 * 197).
The terms \fast and \slow for Systems 1 and 2 were subsequently popularized by~\citet{kahneman2011thinking}.
There is now a growing interest in utilizing a combination of fast and slow reasoning systems in diverse areas of Machine Learning~\citep{anthony_thinking_2017,mujika_fast-slow_2017,schwarzschild_can_2021}.

\begin{figure*}
    \centering
    \includegraphics[scale=0.63]{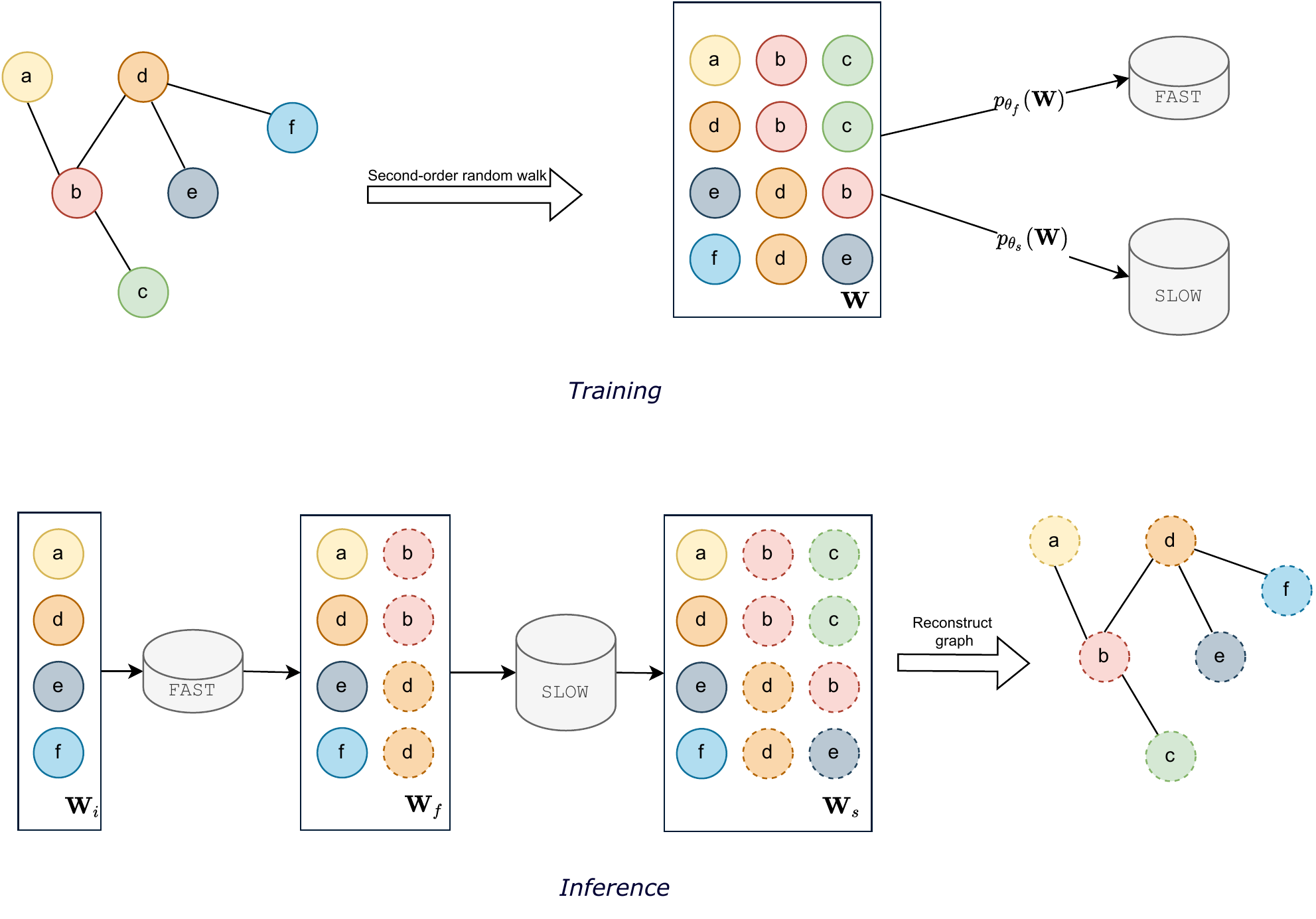}
    \caption{An overview of \ours: During training~(top, \Secref{sec:method}), two auto-regressive models~(\fast and \slow) are trained on a corpus of random walks. The two models have the same architecture, but differ in size (number of parameters). During inference~(below, \Secref{sec:fastslow}), the two models are used in tandem for generating a graph. The \fast model generates the simpler, initial parts of the walk, and the \slow model takes over for generating the latter, more challenging parts.}
    \label{fig:mainfig}
\end{figure*}

This paper introduces \ours, a generative graph model that is inspired by the dual-process theory of mind.
\ours decomposes the problem of generating a graph into the problem of learning to generate walks.
Generating walks provides a setting where identifying the easy and challenging portions is easier: starting from a given node, the model begins by generating walks seen during the training in known neighborhoods.
The difficulty of generating such walks then gradually increases for two reasons.
First, conditioning on increasingly longer contexts is required for generating longer walks.
Second, as the length of the walks exceeds the length seen during training, a model is forced to create neighborhoods not seen during the training: a task that requires more robust generalization capabilities.
\ours leverages this mismatch in problem difficulty by dynamically switching from a small~(\fast) model to a large~(\slow) model for efficient graph generation.
Figure~\ref{fig:mainfig} provides an overview of our approach.
\ours method achieves the same results as using the \slow method alone on three different graphs, while taking up to 50\% less time. 

The backbone of \ours is a decoder-only transformer model, similar to the architectures used by the popular \gptz models.
Using transformers allows us to easily instantiate fast and slow versions of the same model by varying the number of layers.
In contrast to the state-of-the-art methods for generative modeling of graphs that use either an implicit model (\eg GANs as done by \citet{bojchevski2018netgan}), explicit graph distributions~(with no option to vary the parameterization), or generate an entire graph sequence and leverage graph-aware decoding methods~\cite{you2018graphrnn}, our method is simpler~(based on a standard transformer language model) and not sensitive to hyper-parameters~(an identical network setup achieves gains across different graphs.).

\section{\ours}
In this section, we describe our novel graph generation method.
First, we describe how auto-regressive models can be used for graph generation.
Next, we describe how we use two of these models for dynamically for efficient graph generation.

\subsection{Graph generation using auto-regressive models}
\label{sec:method}


\paragraph{Notation}
We denote a graph by \txtgraph.
A random walk \txtrwalk is a sequence of \txtlentrainwalk nodes \rwalkexp obtained by traversing the \txtgraph for \txtlentrainwalk steps starting from $v_1$.
A random walk matrix of \txtntrainwalks such walks is denoted by $\matwalk \in \mathbb{R}^{\ntrainwalks \times \lentrainwalk}$.
An element $\node_{ij} \in \matwalk$ denotes the $j^{th}$ node in thr $i^{th}$ random walk.
For a single random walk \txtrwalk, $\node_i$ denotes the $i^{th}$ node in \txtrwalk.
The \aow{nodes} connected to $\node_i$ are denoted by $\adj(\node_i)$.
We outline the key steps in training and inference (graph generation) below.

\subsubsection{Training}

\paragraph{Step 1: Generating random walks for training} 
Given a graph \txtgraph, we create a second-order random walk matrix $\matwalk \in \mathbb{R}^{\ntrainwalks \times \lentrainwalk}$.
The matrix $\matwalk$ contains $\ntrainwalks$ second-order walks, each of length $\lentrainwalk$.
A second-order random walk~\citep{grover_node2vec_2016} helps in capturing rich topological information of the graph.
Specifically, a node $\node_i$ is sampled as a function of the previous two nodes: $\node_{i-1}$ and $\node_{i-2}$ (and not just $\node_{i-1}$, which will be the case with vanilla sampling).
The details of the sampling procedure are included in Appendix~\ref{sec:secondordersampling}.
Each walk is started by sampling a random node from \txtgraph.

\paragraph{Step 2: Training an auto-regressive model} We use an auto-regressive language model $p_{\theta}$ to learn a generative model of the random walk matrix $p(\matwalk)$.
Specifically, we treat \txtmatwalk as a corpus of $m$ random walks $[\rwalk_1, \rwalk_2, \ldots, \rwalk_{\lentrainwalk}]$ from \txtgraph.
The model is trained to generate the $i^{th}$ node in the walk, conditioned on the preceding~($< i$) nodes.
We model the probability $p(W)$ of a random walk as a series of conditional next token distributions: $ p(\matwalk) =  \prod_{i=1}^{\ntrainwalks} \prod_{j=1}^{\lentrainwalk} p_{\theta} (\node_{ij} \mid \node_{i,<j})$.
 We parameterize $p_\theta$ using a decoder-only language model based on the architecture used by \textsc{gpt-2}~\cite{radford2019language}.
The number of self-attention layers (or \textit{depth}) of the language model decides the number of parameters $\theta$, and, consequently, the strength of the model.


\subsubsection{Inference: graph generation}

\paragraph{Step 3: generating random walks} As the first step of inference, an approximate random walk matrix $\matwalk'$ is obtained by randomly sampling from $p(\matwalk)$. To sample a random walk of length $\lentestwalk$, we first generate a random node $\node_1 \in \graph$.
The generation process begins by $\node_2  \sim p_\theta(v \mid \node_1)$.
The next token is then drawn by sampling $\node_3 \sim p_\theta(v \mid \node_1, \node_2)$.
The process is repeated for \txtlentestwalk -1  steps to generate a random walk of size \txtlentestwalk. 
We generate \txtntestwalks, and stack them to create a generated random walks matrix $\matwalk'$.

 \paragraph{Step 4: Reconstructing graph:}
 We need to assemble the generated graph $\graph'$ from generated random walks $\matwalk'$ generated in the previous step.
We follow the two-step procedure used by~\citet{bojchevski2018netgan} to assemble the generated graph $\graph'$ from generated random walks $\matwalk'$.
First, $\matwalk'$ is converting to a count matrix $\mS$, where $\mS_{ij}$ is the number of times the nodes $\node_i$ and $\node_j$ appeared consecutively~(indicating an edge between $\node_i$ and $\node_j$).
Next, an edge is added between $\node_i$ and $\node_j$ in the generated graph $\graph'$ with probability $p_{ij} = \frac{\mS_{ij}}{\sum_{v \in \adj(i)}\mS_{iv}}$




\paragraph{A note on evaluation} Note that a large model may simply remember a small graph. However, our goal is not such memorization, but rather generalization.
To evaluate this, $\sim 20\%$ of the edges from \txtgraph are hidden during training.
$\graph'$ is then evaluated for presence of these edges.

\paragraph{Relation to language modeling}
Our graph generation method has a 1:1 correspondence with language modeling using graphs.
Our method deals with a graph as characterizing a language, where each random walk \txtmatwalk in \txtgraph is a sentence, and each node \txtnode is a word~(or token).
The language model correspondingly learns to generate valid random walks from \txtgraph. 
Similar ideas were explored by Deepwalk~(\citep{perozzi2014deepwalk}) for learning informative node representations.





\subsection{Fast and slow graph generation}
\label{sec:fastslow}
\newcommand{\lm}{p_{\theta}}

As discussed in the previous section, our method relies on generating random walks.
Let \txtrwalk be a random walk of length \txtlentestwalk to be generated using a trained graph generation model $\lm$, starting from a random node $v_1$.
Since $\lm$ is auto-regressive, the generation process can be succinctly represented using the chain rule.
Let $v_k$ be a node in \txtrwalk with $1 < k < \lentestwalk$.
\begin{align}
    \lm(\rwalk) &= \prod_{i=1}^{k} \lm(v_i \mid v_{< i}) \prod_{j=k+1}^{\lentestwalk} \lm(v_j \mid v_{< j}; v_1, \ldots, v_k)
\end{align}

We posit that there is a $k$ such that the generation of walks $v_1, \ldots, v_k$ and $v_{k+1}, \ldots, v_{\lentestwalk}$ require different levels of difficulty.
Thus, it should be possible to generate the \textit{easy} first part of the walk~($v_1, \ldots, v_k$) using a \fast model, leaving the rest to a \slow model. 

Intuitively, it is easier to generate the first few nodes of a random walk: the first node of the walk is given (the starting point), and generating the second node requires an understanding of a second-order random walk.
Generating subsequent random walks require models to pay attention to the walk seen so far and gets progressively more difficult as the walk length increases.
Further, generating walks longer than \txtlentrainwalk~(random walk length used for training) requires a model with better generalization capabilities.

\paragraph{Instantiating \fast and \slow models} Our  We train two different generation models~(\ie two different $p_{\theta}$) using procedure outlined in \Secref{sec:method}: \fast and \slow.
Both these models have the same architecture type~(transformers), but differ in the number of parameters: \fast is a 1-4 layered transformer whereas \slow has 6 or more layers (depending on the graph).
A speed vs. performance trade-off is expected for the \fast and \slow models: \fast will struggle with generating new walks, whereas \slow will generate these at the cost of slower inference.





Our method, \ours, relies on these key intuitions to pair a fast and slow process together.
We start by generating walks using a \fast model and then switch to a \slow model to explore novel neighborhoods.
Since generation is auto-regressive, such a formulation is natural: subsequent walks can be conditioned on the walks seen so far without any changes to the two models.

\subsection{Switching from \fast to \slow}
\ours proposes to generate the \textit{first} part of the walk quickly using \fast, and the remaining part slowly but more accurately using \slow.
A critical decision for \ours is the \textit{handover} point: at what point should the generation switch from using \fast to \slow?
While generating a walk of length \txtlentestwalk, the switch from \fast to \slow can happen at any point \handover, where $j \in (0, \lentestwalk)$.
However, the choice of \handover is important due to the speed vs. accuracy trade-off: a large $j$ implies that the walk will be generated quickly but mainly using the \fast model.
On the other hand, a smaller $j$ will shift most of the responsibility to the \slow model, for better accuracy but slower inference.
To characterize the difference in performance, we need the notion of a neighborhood, and random walks that perform exploration and exploitation.

\squishlist
\item \textbf{Neighborhood} $\nbr$: a consecutive sequence of \nbrsizetxt nodes that appear in a random walk. For instance, given a random walk $(v_1, v_2, v_3, v_4, v_5)$, and $\nbrsize=4$, the two neighborhoods are $(v_1, v_2, v_3, v_4)$ and $(v_2, v_3, v_4, v_5)$.
\item \textbf{Exploration and exploitation:} a random walk \txtrwalk to be in a state of \textit{exploration} if it is in a neighborhood where it is discovering new neighborhoods not present in the training data.
Otherwise, the walk is said to be in the \textit{exploitation} phase.
As mentioned earlier, a random walk starts from a given node, and thus is expected to be in exploitation mode in the beginning~(known neighborhoods), before switching to exploration mode (new neighborhoods).
Both exploration and exploitation phases are essential: exploration helps the model generalize to new edges, whereas exploitation helps the model recreate the structure.
\squishend

Given these definitions, a sweet spot for the handover point \handover will be the step where the random walk exits the exploration mode and enters the exploitation mode.
To perform this check efficiently, we create a \textit{bloom filter}~\citep{bloom1970space} of all the neighborhoods seen in the training data.

\paragraph{Detecting exploration vs. exploitation}
Given a random walk \txtrwalk, an initial attempt to detect exploration vs. exploitation would be to check if each neighborhood in \txtrwalk is in the training data.
In principle, this can be done by first creating a set of all possible neighborhoods $\sN$ of size $\nbrsize$ in the training data ($\ntrainwalks$ random walks of length $\lentrainwalk$): $\sN = \{(\node_{ij}, \node_{i,j+1}, \ldots, \node_{i, j+p}) \mid i \in [1, \ntrainwalks], j \in [1, \lentrainwalk - \nbrsize + 1]\}$.
Next, a balanced binary tree (available in most programming languages as a hashmap) populated with $\sN$ can be used to efficiently answer membership queries over $\sN$, allowing us to detect exploration vs. exploitation.
In practice, this approach is intractable as the number of all possible $p$ neighborhoods may be exponential.

Using solutions like distributed caching is possible, but may add additional overhead that can cancel any gains obtained using a mix of \fast and \slow models.
Instead, we note that our setup requires a data structure that is less powerful than a hashmap, and allows us to make two concessions: i) we are only interested in checking if a particular neighborhood is absent in the graph, and thus require a reduced set of functions as compared to those supported by a hashmap, and ii) the decision is used to switch to a better (\slow) model, and thus some degree of error is tolerable.
Fortunately, bloom filters exist~\citep{bloom1970space} are widely used for precisely these use cases.

\paragraph{Bloom filter}
A bloom filter $\mathcal{B}$ created over a set $\sS$ provides an efficient way to check if a key $x$ does not exist in $\sS$.
Bloom filters are particularly useful in data-intensive applications, where an application might want to be sure about a query's existence before checking an offline database~\citep{broder2004network,kleppmann2017designing}.

Given a search key $x$, if the search over $\mathcal{B}$ is unsuccessful, it is guaranteed that $x \not \in \sS$.
Otherwise, $x$ may be present with a probability $1 - \probfp$, where $\probfp$ is the false positive rate.
Internally, a bloom filter $\bfilter$ is implemented as an array of $\nbits$ bits accompanied by $\nhashf$ hash functions $\mH_1, \mH_2, \ldots, \mH_\nhashf$.
To add an element $x \in \sS$ to $\bfilter$, each of the $\nhashf$ hash functions map $x$ to $[1, \nbits]$, and thus the corresponding bits are set to 1.
Concretely, $\bfilter[\mH_i(x)] = 1\; \forall i \in [1, \nhashf]$.

To check the presence of an element $x$ in $\bfilter$, it suffices to check if $\exists i \in [1, \nhashf]\; \bfilter[\mH_i(x)] = 0$.
If so, then it is guaranteed that $x \not \in \sS$ (otherwise, all the bits would be set to 1).
Otherwise, the element \textit{may be} present.
Crucially, while creating the bloom filter incurs a one-time cost of $\mathcal{O}(|\sS| \nhashf)$, the lookup can be done in $\mathcal{O}(\nhashf)$ time.
Combined with the space requirements for $\bfilter$, $\nbits << |\sS|$, a bloom filter provides an efficient way to determine if an element is absent from a set.

We use an implementation of scalable bloom filters~\citep{almeida2007scalable}, which are more robust to false positives than the vanilla implementation.
For this implementation, it can be shown that $c \approx M\frac{\log2^2}{|\log \probfp|}$, where $c$ is the capacity, or the maximum number of elements in $\sS$ that a $\bfilter$ with $\nbits$ can support while keeping the false positive rate $\leq \probfp$. 
For completeness, we have included a detailed analysis and relevant algorithms in Appendix~\ref{sec:bloomfilters}.

\paragraph{Bloom filter of neighborhoods}
As noted in \Secref{sec:method}, we generate 100M (second-order) random walks of length 16 for each graph.
We re-use these walks to create a bloom filter $\bfilter$.
For each walk, we use a sliding window of length $\nbrsize = 4$ and inserted the neighborhood in $\bfilter$.
Note that this is a one-time procedure.
Using a false-positive rate of $\probfp=0.01$, the $\bfilter$ is approximately $130\times$ smaller than saving the neighborhoods in a hashmap on average.
Notably, the creation procedure is one-time, and lookup time is a small constant.

Given $\bfilter$, we still need to determine the switching point.
Thus, we sample $50k$ walks using both the \fast and \slow models.
During generation, we query $\mathcal{B}$ with the current neighborhood~(the most recent $\nbrsize$ nodes), and mark the current phase as exploration or exploitation accordingly.

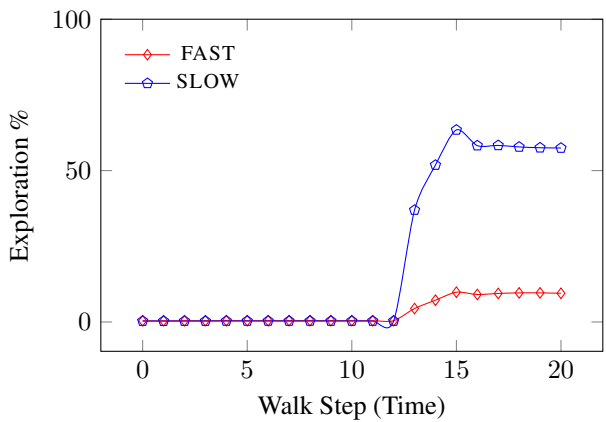
\begin{figure}[!htb]
\begin{tikzpicture}
        \begin{axis}[
        	xlabel=Walk Step (Time),
        	ylabel=Exploration \%,
        	width=1.\columnwidth,height=6cm,
        	ytick={0,50,100},
        	axis background/.style={fill=white},
            legend style={at={(0.03,0.85)},anchor=west, draw=none},
            ymax=100]
        \addplot[color=red,mark=diamond, smooth] coordinates {
(0,0.396)
(1,0.39)
(2,0.416)
(3,0.418)
(4,0.43)
(5,0.418)
(6,0.419)
(7,0.432)
(8,0.441)
(9,0.432)
(10,0.431)
(11,0.418)
(12,0.426)
(13,4.436)
(14,7.21)
(15,9.834)
(16,9.082)
(17,9.425)
(18,9.615)
(19,9.62)
(20,9.478)
        };
        \addplot[color=blue,mark=pentagon, smooth] coordinates {
(0,0.304)
(1,0.285)
(2,0.301)
(3,0.302)
(4,0.323)
(5,0.323)
(6,0.321)
(7,0.312)
(8,0.314)
(9,0.308)
(10,0.317)
(11,0.31)
(12,0.32)
(13,36.966)
(14,51.832)
(15,63.442)
(16,58.252)
(17,58.325)
(18,57.809)
(19,57.553)
(20,57.456)
        };
        \legend{\fast,\slow}
        \end{axis}
        \end{tikzpicture}

\caption{Exploration \% (y-axis) vs. random walk step for \coraml. The larger \slow model explores once the walk exceeds a certain threshold, whereas the lighter \fast model repeats the training data.}
\label{fig:bloomfilter}
\end{figure}

Figure~\ref{fig:bloomfilter} shows for each timestep, and the \% of times the random walk was in exploration mode for both \fast and \slow models.
At the beginning of the walk, the model tends to stick to the same neighborhood (low exploration \%).
The degree of exploration slowly increases as the walk reaches \lentrainwalk.
Then, the model explores new neighborhoods for both \fast and \slow models.
Crucially, note that the extent of exploration is much more significant for the \slow model.
We set the \handoveridx point to be the timestep where the rate of change of exploration is the greatest: $\handoveridx = \argmax_{i} \frac{d EX(i)}{dt}$.
The point is detected using \url{https://pypi.org/project/kneed/}.

In summary, \ours combines \textit{learning} (by training \fast and \slow models) with search~(by using $\bfilter$ to locate optimal handover point) to generate a system that can adapt to the difficulty of the problem for efficient graph generation.

\paragraph{Calculating handover point} We calculate the handover point (the step where we switch from \fast to \slow) for each graph separately. 
We create a bloom filter $\mathcal{B}$ using all the four-node neighborhoods in the training data.
For each graph, we generate 10,000 random walks of length $\lentestwalk = 24$ using both \fast and \slow models.
Then, the handover point is calculated by finding the \textit{knee} of the exploration \% curve, and we use \citet{satopaa2011finding} to find such points.\footnote{\url{https://pypi.org/project/kneed/}}.
We plot the \% of neighborhoods not found in $\mathcal{B}$ (or exploration \%) in Figure~\ref{fig:bloomfilter} for \coraml.

For all the graphs, the \fast model does little exploration.
Notably, the effect is more pronounced for larger graph \polblogs, which proves to be especially challenging for the \fast model~(Figure~\ref{fig:bloomfilterall} in Appendix).

We also experiment with using entropy for deciding the switch, but found it ineffective in determining exploration vs. exploitation Appendix~(\ref{sec:entropyforswitching}), in line with prior work that shows that language models are typically not well-calibrated~\citep{jiang_how_2021}.


\section{Experiments}
\label{sec:experiments}

In this section, we establish the efficacy of our approach with experiments.
First, we show that autoregressive models of graphs can be learned successfully with language models.
Next, we present the results from experiments with \fast and \slow modeling.

\paragraph{Graphs}
We experiment with four representative large graphs: graphs formed by citation networks (\citeseer~\citep{sen2008collective}, \coraml~\citep{mccallum_automating_nodate}), political blogs~(\polblogs~\citep{adamic2005political}, and citation-network for medical publications related to diabetes~(\pubmed~\citep{sen2008collective})) on which implicit graph generation models are shown to perform well.

Graph statistics are provided in Table~\ref{tab:datasets}.
For the link prediction experiments, we use the train/test/val splits provided by \citet{bojchevski2018netgan}.

\begin{table}[ht!]
\centering
\small
\begin{tabular}{@{}lllll@{}}
\toprule
          & \coraml & \citeseer & \polblogs & \pubmed \\ \midrule
$N_{LCC}$ & 2,810   & 2,110     & 1,222     & 19,717  \\
$E_{LCC}$ & 7,981   & 3,757     & 16,714    & 51,913  \\ \bottomrule
\end{tabular}
\caption{Graphs statistics.  The $N_{LCC}$ and $E_{LCC}$  refer to the number of nodes and edges in the largest connected component. We use the dataset supplied by \citet{bojchevski2018netgan} for all experiments. The results for }
\label{tab:datasets}
\end{table}

\paragraph{Tasks and metrics}
Our goal is to learn a generative model of large graphs.
Following prior work, we focus on two different evaluation measures, focused on measuring the ability of the model to learn graph structure and the ability to generalize.
\squishlist

    \item \textbf{Generalization}: a large model may simply \textit{remember} all the random walks seen during training. Thus, the structural metrics are not sufficient for distinguishing between a model that has learned to generalize and a model that is overfitting to the input graph. We follow prior work and evaluate generalization via a link prediction task as a remedy. 
    During training, about 20\% of the edges from each graph are not included in the training data.
    The reconstructed graph $\gG'$ is evaluated to check if these edges are contained. Intuitively, a model that generalizes over the graph instead of regurgitating the training data will perform better when generating unseen edges.
    Link prediction is evaluated using average precision and \auc score, where we use implementation provided by scikit-learn~\citep{scikit-learn} for calculating \auc score.\footnote{We used implementation at: \url{https://scikit-learn.org/stable/modules/generated/sklearn.metrics.roc_auc_score.html\#sklearn.metrics.roc_auc_score}}.
    Recall that the graph is reconstructed from the generated random walks~(\Secref{sec:method}). $p_{ij}$, the normalized probability of an edge between nodes $i$ and $j$, is estimated from the count matrix and supplied to the $\texttt{roc\_auc\_score}$ function as $y\_pred$.
    \item \textbf{Structure}: to evaluate graph structure, we additionally calculate the topological properties of the graph, including the maximum degree, associativity, triangle count, and power-law exp. A detailed definition of these metrics is provided in \Secref{sec:graphstructmetrics} for completeness.
\squishend

\paragraph{\fast, \slow, and \ours models} We base \ours on a decoder-only transformer architecture. 
Specifically, we use a layered-transformer architecture with stacks of self-attention layers~(\sa).
Each \sa layer comprises a self-attention~\citep{vaswani2017attention}, along with a feed-forward layer and skip connections.
To recall from Section~\ref{sec:fastslow}, our experiments involve three models: \begin{inparaenum} [1.)] \item \slow:  larger model with six layers for all datasets except \pubmed, where is has 36 layers. \item \fast: smaller model with a single layer for all datasets, and has 6 layers for \pubmed, and \item \ours: a combination of \fast and \slow. \fast and \slow models are separately trained, and are combined during inference: the first part of the random walk generation is done with \fast, and the second half with \slow. \end{inparaenum}

Other than using larger \fast and \slow models for \pubmed, we do not perform any hyper-parameter tuning: all the models use the same hyperparameters.
We consider the lack of hyper-parameter tuning a core strength of our approach and a key advantage with respect to the baseline.
We do not perform any hyper-parameter tuning: all the models use the same hyperparameters, and use a single Nvidia 2080-ti for all experiments.

\paragraph{Baselines} Note that the main goal of this work is to show that \fast and \slow models can be combined for effective graph generation. Nonetheless find that \ours is competitive with existing graph-generation methods~(\Secref{sec:canautoregdograph}), notably \netgan~\citep{bojchevski2018netgan}.
For completeness, we also compare with a number of parametric, non-parametric, and graph-specific baselines including degree-corrected stochastic block model~(\textsc{dc-sbm}~\citep{karrer2011stochastic}), degree-centrality based adamic-adar index~(AA index~\citep{adamic2003friends}), variational graph autoencoder~\citep{kipf2016variational}, and Node2Vec~\citep{grover_node2vec_2016}.

\begin{figure*}[!ht]
    \centering
    \includegraphics[scale=0.28]{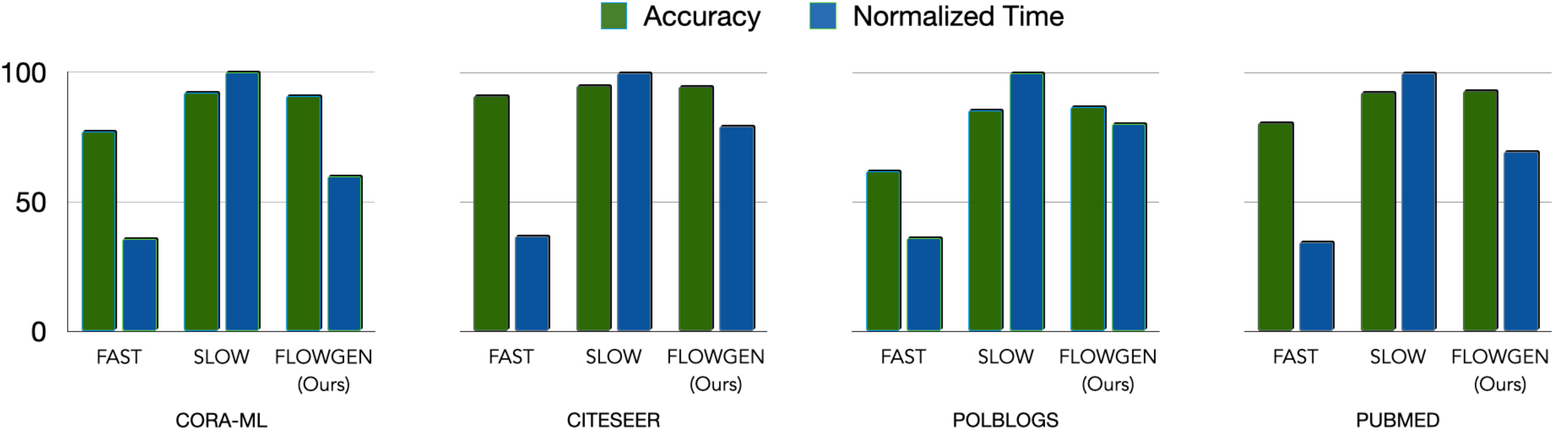}
    \caption{\textbf{Main results:} \auc and time for the different graphs using \fast, \slow, and \ours: \ours is competitive with the larger \slow model, while being upto 2x faster.}
    \label{fig:fastslowresultsbarchart}
\end{figure*}

\subsection{RQ1: Can auto-regressive language models successfully learn generative models of graphs?}
\label{sec:canautoregdograph}
In contrast with prior work, our backbone graph-generation model is a simple transformer-based language model.
The simplicity of this method allows us to experiment with the fast and slow settings easily.
However, does this simplicity come at the cost of performance?
To establish that our graph-generation model is competitive, we evaluate the performance of the larger model, \slow, for link prediction and structural generation for all the graphs.

The results in Table~\ref{tab:coramlstruct} and \ref{tab:linkpredours} show that our transformer-based random walk models achieves competitive performance compared with methods based on either adversarial training or latent variable approaches.
We include additional results on structural prediction in \Secref{sec:appendixexperiments}.
Next, we experiment with \ours, which combines \fast and \slow graphs for generation.

\begin{table}[]
\centering
\small
\begin{tabular}{l|p{2em}p{3em}p{3em}p{3em}p{4.5em}p{4.5em}p{3.5em}p{2.5em}p{2em}}
\toprule
Graph &
  \begin{tabular}[c]{@{}l@{}}Max.\\ degree\end{tabular} &
  \begin{tabular}[c]{@{}l@{}}Assort-\\ ativity\end{tabular} &
  \begin{tabular}[c]{@{}l@{}}Triangle\\ Count\end{tabular} &
  \begin{tabular}[c]{@{}l@{}}Power \\ law exp.\end{tabular} &
  \begin{tabular}[c]{@{}l@{}}Intra-comm \\ unity density\end{tabular} 
  \\ 
  \midrule
CORA-ML    & 240 & -0.075 & 2,814 & 1.860 & 4.3e-4    \\ \midrule
Netgan & \textbf{233} & -0.066 & 1,588 & 1.793 & 6.0e-4   \\\midrule
\fast    &  216   & -0.082     &  \textbf{2,461}     &    \textbf{1.84}   &   5.8e-4        \\ 
\slow    &  200  & \textbf{-0.079}     &  2,143    &    1.853   &   \textbf{5.4e-4}     \\ 
\ours    &  200   & \textbf{-0.080}     &  2,351     &    \textbf{1.84}   &   5.6e-4      \\ 
\bottomrule
\end{tabular}
\caption{Comparison of \slow, \fast, and \ours with Netgan~\citep{bojchevski2018netgan} for structural metrics for \coraml. The ground truth values are listed in the top-row, and the value closer to the ground truth is highlighted in bold. All variants closely match the ground truth graph across a range of metrics.}
\label{tab:coramlstruct}
\end{table}

\begin{table}
\centering
\small
\begin{tabular}{@{}lrrrr}
\toprule
Method & \coraml & \citeseer & \pubmed & \polblogs \\ \midrule
AA-index   & 92.16          & 88.69          & 84.98          & 85.43          \\
\textsc{dc-sbm}        & 96.03          & 94.77          & \textbf{96.76} & 95.46          \\
Node2Vec     & 92.19          & 95.29          & 96.49          & 85.10          \\
\textsc{vgae}        & 95.79          & 95.11          & 94.50          & 93.73          \\
\netgan  & 95.19          & 96.30          & 93.41          & \textbf{95.51} \\
\ours    & \textbf{96.90} & \textbf{96.50} & 93.00          & 93.80          \\ \bottomrule
\end{tabular}
\caption{Comparison of our \slow model with other graph generation baselines on link prediction task. Our graph generation model is competitive. Results for \slow and \fast models are listed in Table~\ref{tab:mainresults}. We find identical trends with average precision and other metrics, results in \Secref{sec:appendixexperiments}, Table~\ref{tab:linkpredallappendix}.}
\label{tab:linkpredours}
\end{table}

\subsection{RQ2: is \ours effective for graph-generation?}
\label{sec:fastslowexp}


Instead of using a fixed handover point, we can also switch dynamically at each step.
However, we found that constantly switching between models incurs a cost as the model has to perform a forward pass on all the tokens seen so far.
This is required, as the auto-regressive attention at each step depends on the hidden layer representations for all layers and previous steps.
A static handover point avoids constant switching and does not degrade the performance.

\paragraph{Results} The results are shown in Table~\ref{tab:mainresults} and Figure~\ref{fig:fastslowresultsbarchart}.
While \slow model outperforms \ours marginally on \coraml and \citeseer, the trade-off is clear from Table~\ref{tab:mainresults}: \ours take considerably less time to achieve similar or better accuracy.
The size of the underlying graph also plays a role in how significant the gains are from our approach: \ours outperforms the \slow model for the large \polblogs graphs.
In contrast, the \fast model is competitive for a smaller graph like \citeseer.
We include additional results in \Secref{sec:appendixexperiments}.

\newcommand{\largehl}[1]{\colorbox{pink}{#1}}
\newcommand{\medhl}[1]{\colorbox{yellow}{#1}}
\newcommand{\smallhl}[1]{\colorbox{green}{#1}}

\begin{table}[]
\centering
\small
\begin{tabular}{@{}lllllll@{}}
\toprule
          & \multicolumn{2}{c}{\fast} & \multicolumn{2}{c}{\slow} & \multicolumn{2}{c}{\ours} \\ \midrule
          & \auc & Time       & \auc & Time    & \auc & Time    \\ \midrule
\coraml   & 91.5        & \smallhl{50k}         & 96.7            & \largehl{180k}    & \textbf{96.9}    & \medhl{110k}   \\
\citeseer & 96.1        & \smallhl{62k}         & \textbf{96.8}   & \largehl{172k}    & 96.5             & \medhl{137k}   \\
\pubmed   & 80.5        & \smallhl{253k}        &  92.1   & \largehl{735k}    & \textbf{93.0}    & \medhl{509k}   \\ 
\polblogs & 66.2        & \smallhl{48k}         & \textbf{93.8}   & \largehl{156k}    & \textbf{93.8}    & \medhl{108k}   \\
\bottomrule
\end{tabular}
\caption{\auc ($\uparrow$) for \fast, \slow, and \ours. The Time (seconds, $\downarrow$) taken by each setup is in parentheses. \ours closely matches or outperforms the larger model \slow while taking a fraction of time.}
\label{tab:mainresults}
\end{table}

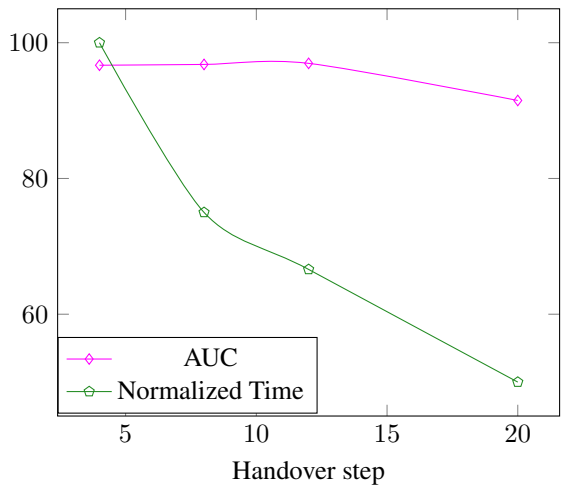
\begin{figure}[!htb]
\begin{tikzpicture}
        \begin{axis}[
        	xlabel=Handover step,
        	width=1\columnwidth,height=7cm,
        	axis background/.style={fill=white},
             legend style={at={(0.0, 0.1)},anchor=west}
            ]
        \addplot[color=Magenta,mark=diamond, smooth] coordinates {
(4,96.68)
(8,96.8)
(12,96.96)
(20,91.49)
        };
        \addplot[color=ForestGreen,mark=pentagon, smooth] coordinates {
(4,100)
(8,75)
(12,66.6)
(20,50)
        };

\legend{AUC, Normalized Time}
\end{axis}
\end{tikzpicture}
\caption{AUC and Normalized time for differenct choices of handover step. When handover to \slow model happens early in the walk (step 4), the time taken is $\sim$ 720 seconds for generating 500 walks, at AUC of $\sim$ 97\%.  Delaying the switch to step 20 leads to a 2x reduction in time taken to generate the walk (360 seconds), with a considerably reduced AUC of $91\%$. \ours offers a tradeoff by calculating the optimal switching point.}
\label{fig:switchingpoint}
\end{figure}

\paragraph{Selection of handover point}
We use a fixed switching point of $13$ for all the graphs. Is this a key design choice? Will delaying the switching point lead to more accurate graphs that are generated slowly? While overall results show that is indeed the case, we conduct a fine-grained analysis of switching point choice for \coraml. The results are shown in Figure~\ref{fig:switchingpoint}. We find that selection of handover point is indeed important.

\paragraph{Performance of \ours with scale}
How does the performance of \ours change as the scale of data increases? We show in \Secref{sec:perfwithscal} that \ours matches or outperforms \slow consistently as the number of walks is increased from 500k to 100m (used current experiments).

\section{Related Work}

\paragraph{Graph generation}
Our work relies on using random walks for learning generative models of graph, similar to~\cite{bojchevski2018netgan} and \cite{you2018graphrnn}.
\cite{you2018graphrnn} learn a generative model of molecules, where each inference step generates the complete graph.
Their setup also leverages graph-aware specialized decoding procedures, and scales for their setup since molecular graphs are typically small.
In contrast, our random walk based method allows learning generative models of large graphs that cannot be generated in a single inference step.
Additionally, in contrast with \cite{bojchevski2018netgan} that use GAN-based training, we leverage relatively simple graph generation model.
\aow{The idea of modeling random walks as sequence of nodes is identical to DeepWalk~\citep{perozzi2014deepwalk}.
However, different from DeepWalk, our main goal is generative graph modeling, and not learning node representations.
Further, our underlying architecture~(transformers) is also different than the one used by DeepWalk (MLP).}


\paragraph{Fast and slow machine learning} There are several works that use the fast-slow metaphor.
For instance, \citet{mujika_fast-slow_2017} present a hierarchical RNN architecture, where the lower (or fast) layer contains one RNN cell for each time-step.
The higher layer in contrast connects several different neurons together.
\citet{hill_grounded_2020} focus on language reasoning tasks, where slow and fast denote the two phases of learning: slow supervised training, and a fast k-shot adaptation.

\aow{Our work is closest in spirit to the remarkable recent work by~\citet{schwarzschild_can_2021,schwarzschild_datasets_2021}, who focus on three different generalization tasks.
They observe increasing the number of test iterations (which corresponds to the network depth in their setting) helps the models in generalizing better to the difficult problem.
Our study replicates this general finding, by showing that \fast (small) and \slow (larger) models can be combined for efficient graph generation.
Our method can be seen as an extension of their method for graph generation, with the following novel additions.
First, instead of varying the depth of the network, we actually leverage two different transformer networks (\fast and \slow), and the output of \fast is used by \slow.
Second, we determine the switching point in a principled fashion using bloom filters.
\citet{schwarzschild_can_2021} note that the confidence of the model was a good proxy for correctness in their setting.
We find that not to be the case, and also propose a method for finding a switching point for the network.
}

\paragraph{Adaptive computation}
A related body of work on adaptive computation seeks to preempt computation based on intermediate representations~\citep{liu2020fastbert,zhou2020bert,schuster2021consistent,geng2021romebert}.
Different from these methods, our approach completely obviates making any architectural modifications.
As the attached code shows, the \fast and \slow models are initialized identically, with the difference of the number of layers.
The switch from \fast to \slow is also simple: \ours moves intermediate outputs from a \fast to a \slow model at an optimal step, and the auto-regressive nature of our graph generation setup guarantees that the setup remains well-defined.
\citet{schuster2022confident} present \textsc{clam}, a language model that performs language generation adaptively.
In 

\section{Conclusion}
Future machine learning applications will potentially have API-level access to several models of varying strengths and costs of usage. In such scenarios, building systems that can adapt to the difficulty of the sample will be critical for scale and efficiency.
\ours presents a real-world use case for such \fast-\slow systems.
As future work, we plan to explore the use of \fast-\slow generation methods for effective and adaptive language generation using large-language models.



\section*{Acknowledgment}
This material is partly based on research sponsored
in part by the Air Force Research Laboratory under
agreement number FA8750-19-2-0200. The U.S.
Government is authorized to reproduce and distribute reprints for Governmental purposes notwithstanding any copyright notation thereon. The views
and conclusions contained herein are those of the
authors and should not be interpreted as necessarily
representing the official policies or endorsements,
either expressed or implied, of the Air Force Research Laboratory or the U.S. Government.

\newpage
\clearpage
\bibliography{main}
\bibliographystyle{icml2022}
\newpage
\clearpage
\newpage
\clearpage
\appendix

\section{Overview of Bloom Filters}
\label{sec:bloomfilters}

\normalem
\SetKwInput{KwGiven}{Given}
\SetKwInput{KwInit}{Init}
\SetKwInput{KwReturn}{return}

\begin{algorithm}[!htb]
\SetAlgoLined
\KwGiven{$\bfilter$, $\mH$, $\sS$}

\KwInit{$\bfilter(i) \gets 0; i \in [1, M]$}

\For(\tcp*[h]{$\mathcal{O}(|\sS|)$}){$q \in \sS$} {
\For(\tcp*[h]{$\mathcal{O}(|\mH|) = \mathcal{O}(k)$}){$i \gets 1, 2, \ldots, \nhashf$}{
$\bfilter(\mH_i(q)) \gets 1$
}
}
\caption{Creating a bloom filter with $M$ bits and $h$ hash functions $\mH$ over a set $\sS$. Each hash function takes $\mathcal{O}(1)$, and thus creating a bloom filter incurs a one time cost $\mathcal{O}(\nhashf|\sS|)$.}
\label{alg:creatingbloomfilter}
\end{algorithm}

\begin{algorithm}[tb]
\SetAlgoLined
\KwGiven{$\bfilter$, $\mH$}

\For(\tcp*[h]{$\mathcal{O}(\nhashf)$}){$i \gets 1, 2, \ldots, \nhashf$}{
\If(\tcp*[h]{certainly absent}){$\bfilter(\mH_i(q)) = 0 $} { 
\KwRet{False}
}
}
\tcc{Maybe present with a false positive rate $p$.}
\KwRet{True}
\caption{Querying a bloom filter. The cost is a fixed constant $\mathcal{O}(\nhashf)$.}
\label{alg:queyringbloomfilter}
\end{algorithm}

A bloom filter \bfilterm over a set $\sS$ is a data structure for efficient set-membership queries.
The time to search is independent of the number of elements in $\sS$.
As a trade-off, a bloom filter can generate false positives (indicate that a query $q \in \sS$ when it is absent).
We will return to an analysis of false-positive rate after expanding on details of a bloom filter.

Given a search key $x$, if the search over $\mathcal{B}$ is unsuccessful, it is guaranteed that $x \not \in \sS$.
Otherwise, $x$ may be present with a probability $1 - \probfp$, where $\probfp$ is the false positive rate.
Internally, a bloom filter $\bfilter$ is implemented as an array of $\nbits$ bits accompanied by $\nhashf$ hash functions $\mH_1, \mH_2, \ldots, \mH_\nhashf$.
To add an element $x \in \sS$ to $\bfilter$, each of the $\nhashf$ hash functions map $x$ to $[1, \nbits]$, and thus the corresponding bits are set to 1.
Concretely, $\bfilter[\mH_i(x)] = 1\; \forall i \in [1, \nhashf]$.

To check the presence of an element $x$ in $\bfilter$, it suffices to check if $\exists i \in [1, \nhashf]\; \bfilter[\mH_i(x)] = 0$.
If so, then it is guaranteed that $x \not \in \sS$ (otherwise, all the bits would be set to 1).
Otherwise, the element \textit{may be} present.
Crucially, while creating the bloom filter incurs a one-time cost of $\mathcal{O}(|\sS| \nhashf)$, the lookup can be done in $\mathcal{O}(\nhashf)$ time.
Combined with the space requirements for $\bfilter$, $\nbits << |\sS|$, a bloom filter provides an efficient way to determine if an element is absent from a set.
r
The key elements in the design of a bloom filter are its size $M$, $\nhashf$ hash functions $\mH_1, \mH_2, \ldots, \mH_{\nhashf}$, and the size of set $\sS$ over which search operations are to be performed.

Algorithms~\ref{alg:creatingbloomfilter} and \ref{alg:queyringbloomfilter} show the algorithms for creating and querying a bloom filter, respectively.

\begin{table*}
\centering
\begin{tabular}{@{}llll@{}}
\toprule
          & \fast & \slow  &    \ours \\ \midrule
\coraml~(500k)   & 76.8~(288) &  \textbf{92.2}~(806) &  90.8~(484) \\
\coraml~(100M)   & 91.5~(50k) &  96.7~(180k) &  \textbf{96.9}~(110k) \\
\citeseer~(500k) & 90.9~(313) &  \textbf{94.6}~(862) &  93.3~(687)  \\
\citeseer~(100M) & 96.1~(62k) & \textbf{96.8}~(172k) &  96.5~(137k) \\
\polblogs~(500k) & 61.9~(309) & 85.4~(854) & \textbf{86.5}~(686) \\
\polblogs~(100M) & 66.2~(48k) & \textbf{93.8}~(156k) & \textbf{93.8}~(108k) \\
\pubmed~(500k) & 58.36~(1200) & 71.04~(854) & \textbf{71.18}~(686) \\
\pubmed~(100M) & 80.53~(253k)  & \textbf{92.09}~(735k) & \textbf{92.97}~(509k) \\ \bottomrule
\end{tabular}
\caption{\textbf{Main results:} AUC for \fast, \slow, and \ours, a combination of \fast-\slow models. The time (seconds) taken by each setup is in parentheses. \ours closely matches or outperforms the larger model \slow while taking a fraction of time.}
\label{tab:fastslowapwithscale}
\end{table*}

One of the biggest follies of a bloom filter are its false positive rates.
\citet{chang2004approximate} proposed bucketed bloom filters to alleviate the false positive rate.
In their method, each hash function $\mH_i$ maps to the indices $[(i - 1) * m + 1, m]$, where $m = M / \nhashf$ is the number of bits in each bucket.

Let $P$ be the rate of false positives, $|\sS| = n$.
Allowing each bucket of bloom filter to be 50\% full, it can be shown that the number of elements $n \sim M \frac{(ln2)^2}{|lnP|}$~\citep{almeida2007scalable}.
See~\citet{christensen2010new} for a comprehensive analysis of false positive rate for classical implementation of bloom filters.

We next approximate the size of bloom filter required for storing all neighborhoods of a graph $\gG$.
Let $|\nodeset|$ be the number of nodes in $\gG$.
Let $\maxdegree$ be the max-degree of $\gG$.
Then, the number of neighborhoods $\nbr$ of size $\nbrsize$ are upper-bounded by $|\nodeset| * \maxdegree^{\nbrsize - 1}$.
Clearly, this can be non-tractable for large, dense graphs.
However, if $\maxdegree$ is a fixed constant, then the number of neighborhoods is $\mathcal{O}(|\nodeset|)$~($ \maxdegree^{\nbrsize - 1}$ is absorbed in the constant).
Thus, for such graphs, bloom filter can be tractably grown.
Crucially, note that our goal is not to store all the graphs.
Rather, we want to only approximately answer the membership queries in the graph.

\begin{figure*}[!htb]
\minipage{0.33\textwidth}
\begin{tikzpicture}
        \begin{axis}[
        	xlabel=Walk Step (Time),
        	ylabel=Exploration \%,
        	width=1.\columnwidth,height=6cm,
        	ytick={0,50,100},
        	axis background/.style={fill=white},
            legend style={at={(0.03,0.85)},anchor=west, draw=none},
            ymax=100]
        \addplot[color=red,mark=diamond, smooth] coordinates {
(0,0.396)
(1,0.39)
(2,0.416)
(3,0.418)
(4,0.43)
(5,0.418)
(6,0.419)
(7,0.432)
(8,0.441)
(9,0.432)
(10,0.431)
(11,0.418)
(12,0.426)
(13,4.436)
(14,7.21)
(15,9.834)
(16,9.082)
(17,9.425)
(18,9.615)
(19,9.62)
(20,9.478)
        };
        \addplot[color=blue,mark=pentagon, smooth] coordinates {
(0,0.304)
(1,0.285)
(2,0.301)
(3,0.302)
(4,0.323)
(5,0.323)
(6,0.321)
(7,0.312)
(8,0.314)
(9,0.308)
(10,0.317)
(11,0.31)
(12,0.32)
(13,36.966)
(14,51.832)
(15,63.442)
(16,58.252)
(17,58.325)
(18,57.809)
(19,57.553)
(20,57.456)
        };
        \legend{\fast,\slow}
        \end{axis}
        \end{tikzpicture}
\endminipage\hfill
\minipage{0.33\textwidth}
\begin{tikzpicture}
        \begin{axis}[
        	width=1.15\columnwidth,height=6cm,
        	ytick={0,50,100},
        	axis background/.style={fill=white},
            ymax=100]
        \addplot[color=red,mark=diamond, smooth] coordinates {
(0,0.18)
(1,0.186)
(2,0.193)
(3,0.2)
(4,0.207)
(5,0.195)
(6,0.189)
(7,0.189)
(8,0.198)
(9,0.203)
(10,0.204)
(11,0.204)
(12,0.203)
(13,27.938)
(14,37.809)
(15,46.143)
(16,36.987)
(17,36.298)
(18,35.748)
(19,35.628)
(20,35.649)
        };
        \addplot[color=blue,mark=pentagon, smooth] coordinates {
(0,0.125)
(1,0.138)
(2,0.143)
(3,0.145)
(4,0.145)
(5,0.14)
(6,0.139)
(7,0.14)
(8,0.137)
(9,0.137)
(10,0.135)
(11,0.142)
(12,0.141)
(13,48.477)
(14,63.163)
(15,73.061)
(16,69.712)
(17,69.958)
(18,69.999)
(19,70.056)
(20,69.846)
        };
        \end{axis}
        \end{tikzpicture}
\endminipage\hfill
\minipage{0.33\textwidth}%
\begin{tikzpicture}
        \begin{axis}[
        	width=1.15\columnwidth,height=6cm,
        	axis background/.style={fill=white},
        	ytick={0,50,100},
            ymax=100]
        \addplot[color=red,mark=diamond, smooth] coordinates {
(0,12.552)
(1,13.365)
(2,14.852)
(3,14.601)
(4,15.262)
(5,14.999)
(6,15.589)
(7,15.526)
(8,15.547)
(9,15.442)
(10,15.671)
(11,15.695)
(12,15.722)
(13,15.666)
(14,16.007)
(15,16.018)
(16,15.78)
(17,15.693)
(18,15.481)
(19,15.783)
(20,15.75)
        };
        \addplot[color=blue,mark=pentagon, smooth] coordinates {
(0,16.818)
(1,17.416)
(2,19.827)
(3,19.528)
(4,20.351)
(5,20.244)
(6,20.522)
(7,20.343)
(8,20.64)
(9,20.623)
(10,20.721)
(11,20.773)
(12,20.867)
(13,27.426)
(14,36.769)
(15,43.791)
(16,43.386)
(17,43.949)
(18,43.818)
(19,43.924)
(20,43.737)
        };
        \end{axis}
        \end{tikzpicture}
\endminipage
\caption{Exploration \% (y-axis) vs. random walk step for \coraml~(left), \citeseer~(middle), and \polblogs~(right). For all the graphs, the larger \slow model explores once the walk exceeds a certain threshold, whereas the lighter \fast model repeats the training data.}
\label{fig:bloomfilterall}
\end{figure*}
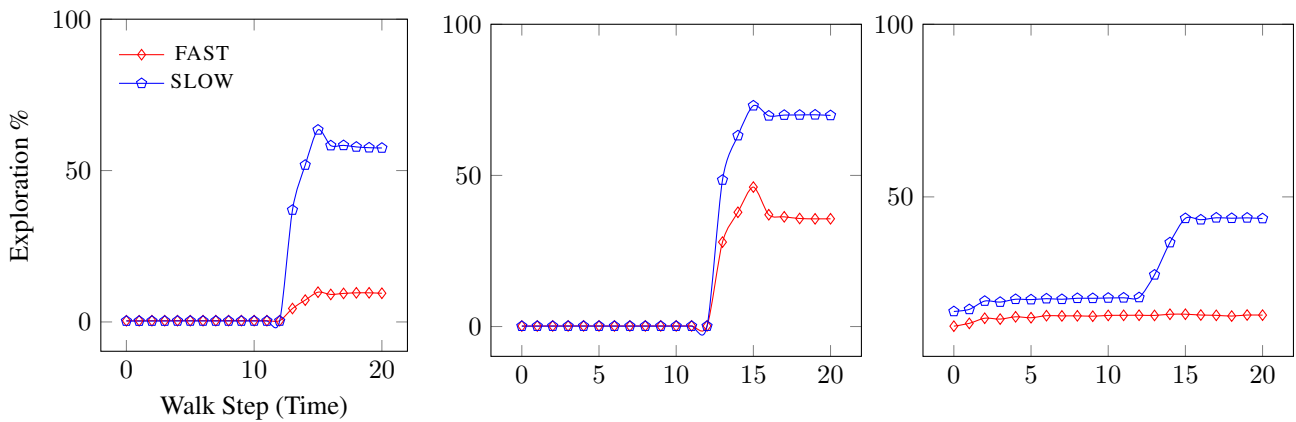


\section{Second-order sampling for generating the training data}
\label{sec:secondordersampling}
\aow{For completeness, we now present the second order sampling method used by~\citet{grover_node2vec_2016} that we adopt for generating the training data for our system.}

\aow{Following the notation used by \citet{grover_node2vec_2016}, let $t$ be the previous node visited by the walk, and $v$ be the current node (\ie the walk just traversed $[t, v]$).
The distribution over the next node $x$, $\pcond{x}{t, v}$, is given as $\pcond{x}{t, v} = \frac{\pi(x, t)}{\sum_{y \in \adj(v)}\pi(y, t)}$.
Here, $\pi(x, t)$ is defined as follows:}

\begin{equation*}
\pi(x, t) = \begin{cases}
\frac{1}{p} &\text{if $d_{tx} = 0$}\\
1 &\text{if $d_{tx} = 1$}\\
\frac{1}{q} &\text{if $d_{tx} = 2$}
\end{cases}
\end{equation*}

\aow{The parameter $p$ decides the likelihood of revisiting a node. Specifically, a low $p$ will encourage the walk to go back to the node $t$ recently visited.
Similarly, $q$ controls the likelihood of the walk visiting new nodes.
A lower value of $q$ will encourage the walk to move towards node that are farther away from the node recently visited, allowing higher exploration.
Following~\citet{bojchevski2018netgan}, we set $p=q=1$ to balance between the two properties.
For more insights into the properties of second order random walk, please see Section 3.2 of~\citep{grover_node2vec_2016}.
}

\section{Additional Results and Experimental Setup}
\label{sec:appendixexperiments}

\paragraph{Experimental Setup} All the models were trained using a single Nvidia 2080-Ti GPU. During inference, we were able to fit both the models on a single GPU. We found that storing the models on separate GPUs erases some of the gains of \ours, due to required data transfer across machines.
Implementation is done in PyTorch Lightning.\footnote{\url{https://www.pytorchlightning.ai/}}.
Implementation of a number of evaluation and data generation scripts was derived from open-source implementation of \citet{bojchevski2018netgan}.\footnote{\url{https://github.com/danielzuegner/netgan}}

\begin{figure*}
    \centering
    \includegraphics[scale=0.25]{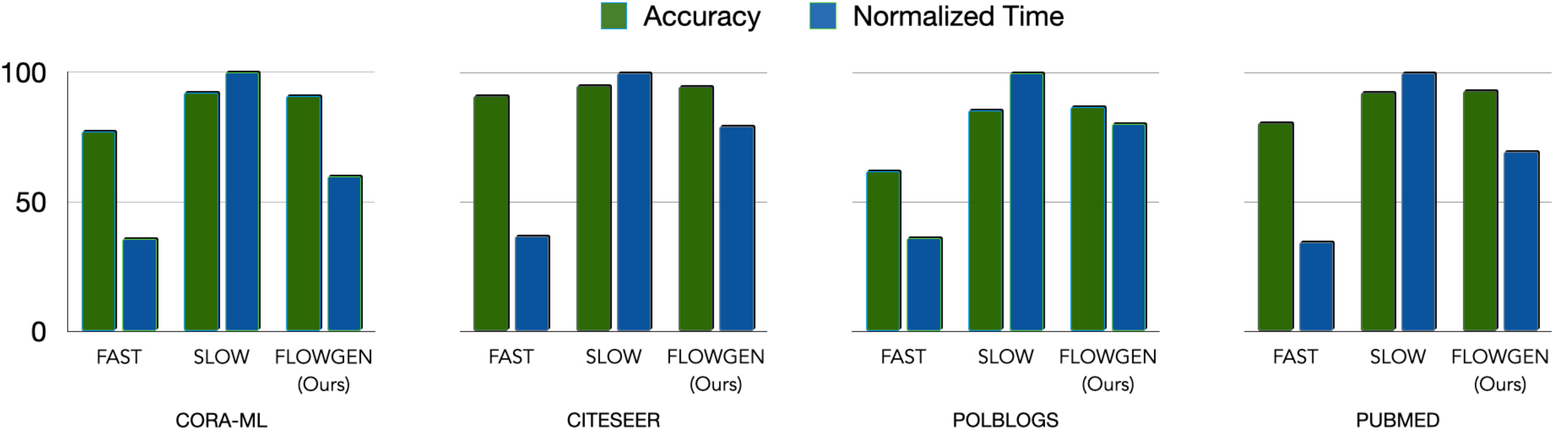}
    \caption{Average precision vs. time taken for the three graphs. The \fast and \slow model speed-accuracy trade-off is apparent: \fast model is fast but less accurate (average precision $\sim$ 75\%, compared to the \slow model which is slower but has average precision of 92\%. \fastslow combines the strengths of the two modes: it achieves an accuracy of 90\% while being $\sim$ 50\% faster than the \slow model. Note that the time is normalized relative to \slow (\slow takes 100\% of the time).}
    \label{fig:fastslowaptimeall}
\end{figure*}

\subsection{Graph structure metrics}
\label{sec:graphstructmetrics}

Table~\ref{tab:structcomplete} shows the structural metrics for all the graphs. For the mechanism to calculates these metrics, please see Appendix A of \citet{bojchevski2018netgan}. Here, we instead provide an alternate and informal, high-level overview of each metric to help with interpretation of Table~\ref{tab:structcomplete}.
\begin{enumerate}
    \item \textbf{Max. degree}: maximum degree across all nodes. Used to approximate the degree of density of the generated graph.
    \item \textbf{Assortativity}: pearson correlation of degrees of connected nodes. Similar values for two different graphs indicates a similarity in topology.
    \item \textbf{Triangle count}: number of triangles in a graph (set of three vertices connected to each other).
    \item \textbf{Intra/Inter density}: fraction of edges that are part of the same community/fraction of edges that cross communities.
    \item \textbf{Charac. path len} (characteristic path length): number of edges in the shortest path between any two vertices.
    \item \textbf{Clustering coefficient}: For a given node $v$, let $\mathcal{N}(v)$ be its set of neighbors. Informally, clustering coefficient is the ratio of number of edges that exist within $\mathcal{N}(v)$, to the number of edges that can possibly exist within $\mathcal{N}(v)$.
\end{enumerate}

\begin{table*}[!ht]
\centering
\small
\begin{tabular}{llp{2em}p{4em}p{3em}p{3em}p{4.5em}p{4.5em}p{3.5em}p{3em}}
\toprule
Graph & Type & 
  \begin{tabular}[c]{@{}l@{}}Max.\\ degree\end{tabular} &
  \begin{tabular}[c]{@{}l@{}}Assort-\\ ativity\end{tabular} &
  \begin{tabular}[c]{@{}l@{}}Triangle\\ Count\end{tabular} &
  \begin{tabular}[c]{@{}l@{}}Power \\ law exp.\end{tabular} &
  \begin{tabular}[c]{@{}l@{}}Inter-comm \\ unity density\end{tabular} &
  \begin{tabular}[c]{@{}l@{}}Intra-comm.\\ unity denisty\end{tabular} &
  \begin{tabular}[c]{@{}l@{}}Cluster-\\ ing coeff.\end{tabular} &
  \begin{tabular}[c]{@{}l@{}}Charac.\\ path len.\end{tabular}  \\ \midrule
\coraml & \fast & 216.0 & -0.08186 & 2461 & 1.84745 & 0.00129 & 0.00058 & 0.00317 & 5.59302\\
\coraml & \slow & 200.0 & -0.07949 & 2143 & 1.84531 & 0.0013 & 0.00055 & 0.00333 & 5.40631\\
\coraml & \ours & 200.0   & -0.080     &  2,351     &1.84 & 0.00129 & 0.00056 & 0.00395 & 5.50565\\ \midrule
\citeseer & \fast & 59.0 & -0.04444 & 427 & 2.19731 & 0.00114 & 0.00025 & 0.01601 & 9.73914\\
\citeseer & \slow & 55.0 & -0.04329 & 437 & 2.18366 & 0.00114 & 0.00026 & 0.0195 & 9.80901\\
\citeseer & \ours & 61.0 & -0.03117 & 455 & 2.17116 & 0.00116 & 0.00025 & 0.01872 & 9.87617\\
\midrule
\polblogs & \fast & 254.0 & -0.30609 & 44428 & 1.43388 & 0.00534 & 0.01406 & 0.00438 & 2.85113\\
\polblogs & \slow & 273.0 & -0.23833 & 52742 & 1.43739 & 0.0054 & 0.014 & 0.0047 & 2.80242\\
\polblogs & \ours & 289.0 & -0.25944 & 49204 & 1.43222 & 0.00541 & 0.01397 & 0.00442 & 2.79202\\ \midrule
\pubmed & \fast & 115.0 & -0.1228 & 4970 & 2.29702 & 4e-05 & 0.00015 & 0.00348 & 6.84598\\
\pubmed & \slow & 106.0 & -0.14983 & 4089 & 2.2487 & 4e-05 & 0.00015 & 0.00321 & 6.76372\\
\pubmed & \ours & 111.0 & -0.14689 & 4172 & 2.25055 & 4e-05 & 0.00015 & 0.00324 & 6.76702\\

\bottomrule
\end{tabular}
\caption{Structural metrics for all graphs used in this work. \ours closely matches \slow, but takes only a fraction of time.}
\label{tab:structcomplete}
\end{table*}

\begin{table*}
\centering
\small
\begin{tabular}{@{}lllllllll@{}}
\toprule
Method       & \multicolumn{2}{c}{\coraml}     & \multicolumn{2}{c}{\citeseer}   & \multicolumn{2}{c}{\pubmed} & \multicolumn{2}{c}{\polblogs} \\ \midrule
              & \auc  & \ap   & \auc  & \ap   & \auc           & \ap            & \auc           & \ap   \\ \midrule
Adamic/Adar   & 92.16 & 85.43 & 88.69 & 77.82 & 84.98          & 70.14          & 85.43          & 92.16 \\
DC-SBM        & 96.03 & 95.15 & 94.77 & 93.13 & \textbf{96.76} & 95.64          & 95.46          & 94.93 \\
node2vec      & 92.19 & 91.76 & 95.29 & 94.58 & 96.49          & 95.97          & 85.10          & 83.54 \\
VGAE          & 95.79 & 96.30 & 95.11 & 96.31 & 94.50          & \textbf{96.00} & 93.73          & 94.12 \\ \midrule
NetGAN (500K) & 94.00 & 92.32 & 95.18 & 91.93 & 87.39          & 76.55          & 95.06          & 94.61 \\
NetGAN (100M) & 95.19 & 95.24 & 96.30 & 96.89 & 93.41          & 94.59          & \textbf{95.51} & 94.83 \\
\ours (100M) & \textbf{96.93} & \textbf{97.22} & \textbf{96.8} & \textbf{97.45} & 93.0       & 91.16        & 93.8    & \textbf{95.05}    \\

\bottomrule
\end{tabular}
\caption{Comparison of \ours with baselines on link prediction task for six different graphs.}
\label{tab:linkpredallappendix}
\end{table*}

\subsection{Performance of \ours with scale}
\label{sec:perfwithscal}

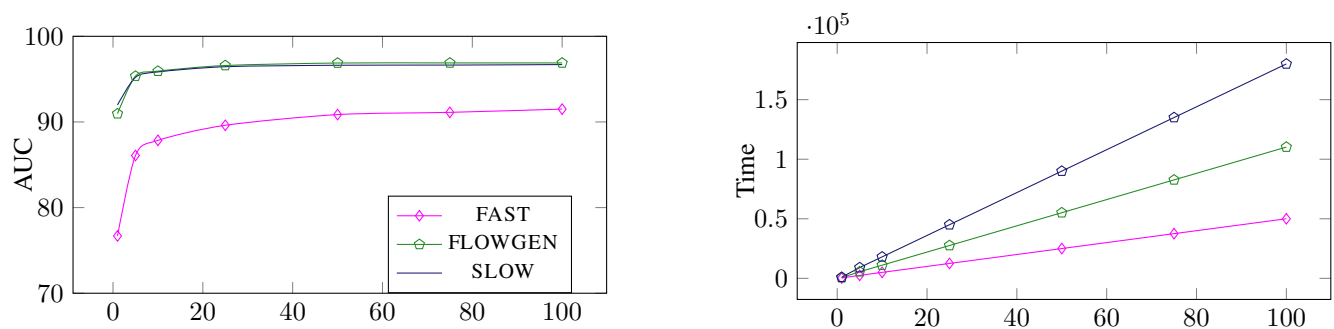
\begin{figure*}[!htb]
\minipage{0.44\textwidth}
\begin{tikzpicture}
        \begin{axis}[
        	ylabel=AUC, 
        	width=1.15\columnwidth,height=5cm,
        	axis background/.style={fill=white},
            ylabel near ticks, ylabel shift={-10pt},
            legend style={at={(0.59,0.19)},anchor=west},
            ymin=70,
            ymax=100]
        \addplot[color=Magenta,mark=diamond, smooth] coordinates {
(1, 76.69)
(5, 86.08)
(10, 87.85)
(25, 89.61)
(50, 90.86)
(75, 91.12)
(100, 91.5)
        };
        \addplot[color=ForestGreen,mark=pentagon, smooth] coordinates {
(1, 90.96)
(5, 95.35)
(10, 95.93)
(25, 96.59)
(50, 96.88)
(75, 96.89)
(100, 96.89)
        };
        \addplot[color=MidnightBlue,mark=circle, smooth] coordinates {
(1, 91.98)
(5, 95.21)
(10, 95.84)
(25, 96.45)
(50, 96.62)
(75, 96.64)
(100, 96.7)
        };

\legend{\fast, \ours, \slow}
\end{axis}
\end{tikzpicture}
\endminipage\hfill
\minipage{0.44\textwidth}
\begin{tikzpicture}
        \begin{axis}[
        	ylabel=Time, 
        	width=1.15\columnwidth,height=5cm,
        	axis background/.style={fill=white},
        ylabel near ticks, ylabel shift={-7pt},
            ]
        \addplot[color=Magenta,mark=diamond, smooth] coordinates {
(1,250)
(5,2500)
(10,5000)
(25,12500)
(50,25000)
(75,37500)
(100,50000)
        };
        \addplot[color=ForestGreen,mark=pentagon, smooth] coordinates {
(1,551)
(5,5510)
(10,11020)
(25,27550)
(50,55100)
(75,82650)
(100,110200)
        };
        \addplot[color=MidnightBlue,mark=pentagon, smooth] coordinates {
(1,900)
(5,9000)
(10,18000)
(25,45000)
(50,90000)
(75,135000)
(100,180000)
        };

\end{axis}
\end{tikzpicture}
\endminipage\hfill

\caption{AUC and time taken~(y-axis) for the three models for \coraml, as the number of random walks sampled increases from 500k to 100M.}
\label{fig:roctimewithscale}
\end{figure*}

How does the performance of \ours change as the scale of data increases? To test this, we vary the number of random walks $\ntestwalks$ generated during inference to recreate the graph.
The results are shown in Figure~\ref{fig:roctimewithscale}.  
\ours matches or outperforms \slow, while being consistently faster across the number of walks.
Table~\ref{tab:performance_across_scale} shows the \auc for different graphs for 500k and 100M walks.

\begin{table*}[!ht]
\centering
\begin{tabular}{@{}lllllll@{}}
\toprule
          & \multicolumn{2}{c}{\fast} & \multicolumn{2}{c}{\slow} & \multicolumn{2}{c}{\ours} \\ \midrule
          & \auc         & Time (s)        & \auc              & Time (s)   & \auc              & Time (s)   \\ \midrule
\coraml~(500k)   & 76.8        & 288         & 92.2             & 806    & 90.8             & 484    \\
\coraml~(100M)   & 91.5        & 50k         & 96.7             & 180k   & \textbf{96.9}    & 110k   \\\midrule
\citeseer~(500k) & 90.1        & 313         & 94.6             & 862    & 93.3             & 687    \\
\citeseer~(100M)  & 96.1        & 62k         & \textbf{96.8}    & 172k   & 96.5             & 137k   \\\midrule
\polblogs~(500k) & 61.9        & 309         & 85.4             & 854    & 86.5             & 686    \\
\polblogs~(100M)  & 66.2        & 48k         & \textbf{93.8}    & 156k   & \textbf{93.8}    & 108k   \\\midrule
\pubmed~(500k)   & 61.9        & 309         & 85.4             & 854    & 86.5             & 686    \\
\pubmed~(100M)    & 80.5        & 253k        & 92.1    & 735k   & \textbf{93.0}    & 509k   \\ \bottomrule
\end{tabular}
\caption{Performance of \slow, \fast, and \ours for different number of sampled random walks: \ours is competitive across scale.}
\label{tab:performance_across_scale}
\end{table*}





\subsection{Using entropy for deciding the switch}
\label{sec:entropyforswitching}
Our method of switching from \fast to \slow model relies on the presence of the walk in training set.
This can be seen 
We also experiment with using entropy for deciding the switch, but found it ineffective in determining exploration vs. exploitation Appendix~(\ref{sec:entropyforswitching}).
\aow{Recall that we are using an auto-regressive language model for generating the walks.
Thus, at each step i, the model generates a distribution over the next node, $\pcond{\node_i}{\node_1, \node_2, \ldots, \node_{i-1}}$.
Thus, for a well calibrated model, in the exploitation phase, when the model is still generating walks from the training set, the entropy of this distribution will be fairly low~(the model will be confident about the next node), and that the entropy will increase further in the walk.
If that was the case, the entropy of the distribution can be a useful indicator of the switching point.
We investigate the same in this section.
}

\begin{figure*}[!htb]
\minipage{0.3\textwidth}
\includegraphics[width=\linewidth]{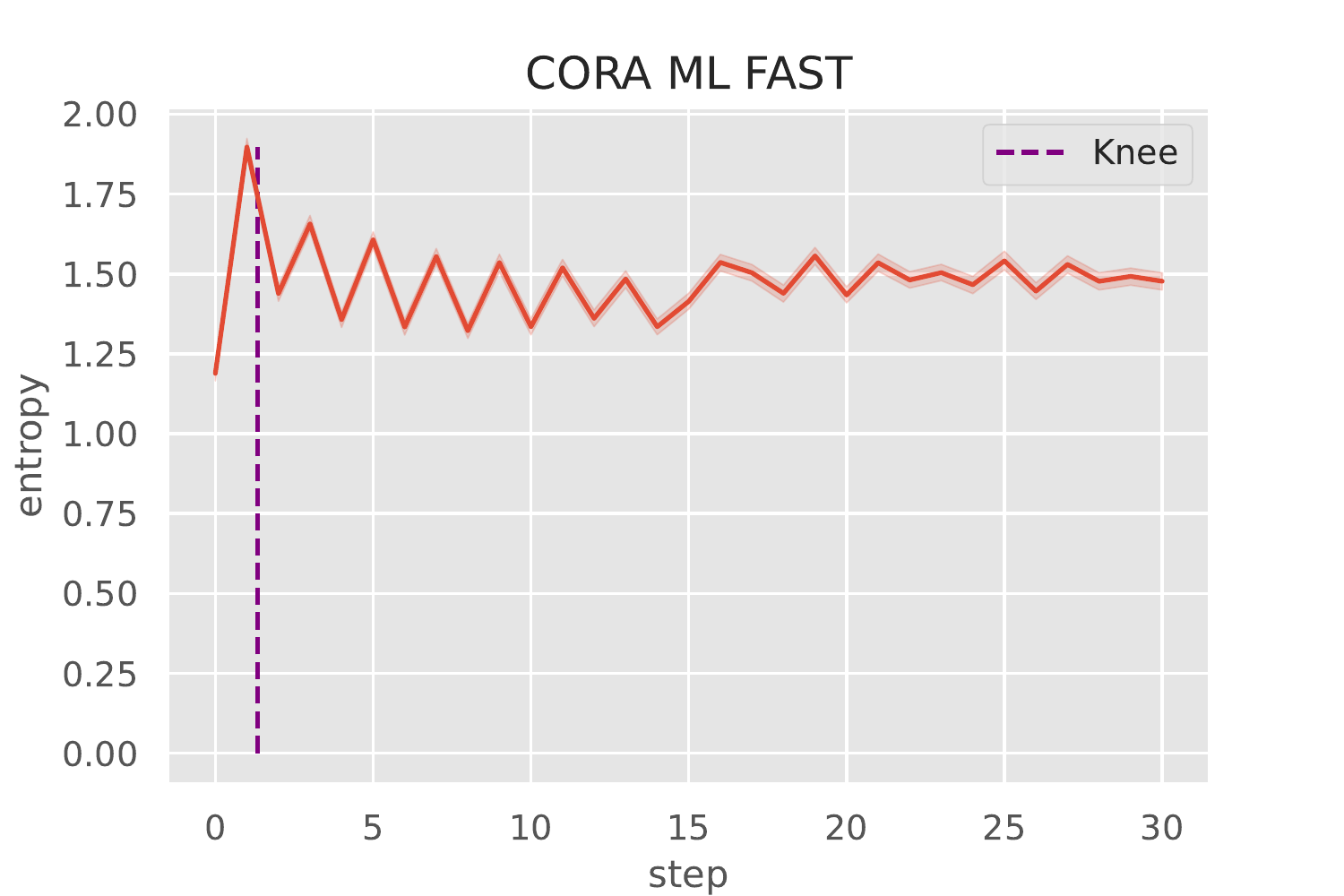}
\endminipage\hfill
\minipage{0.3\textwidth}
\includegraphics[width=\linewidth]{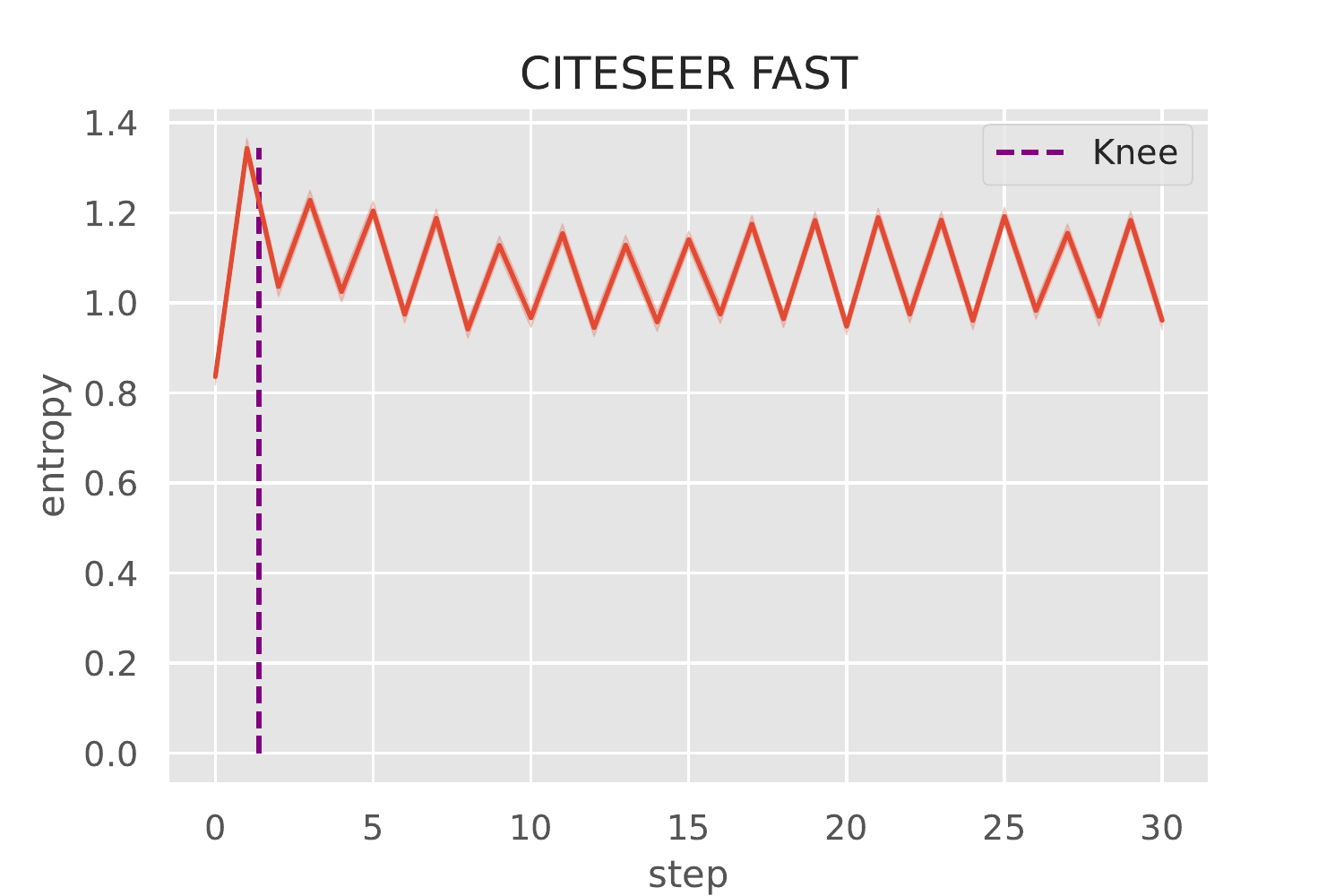}
\endminipage\hfill
\minipage{0.3\textwidth}
\includegraphics[width=\linewidth]{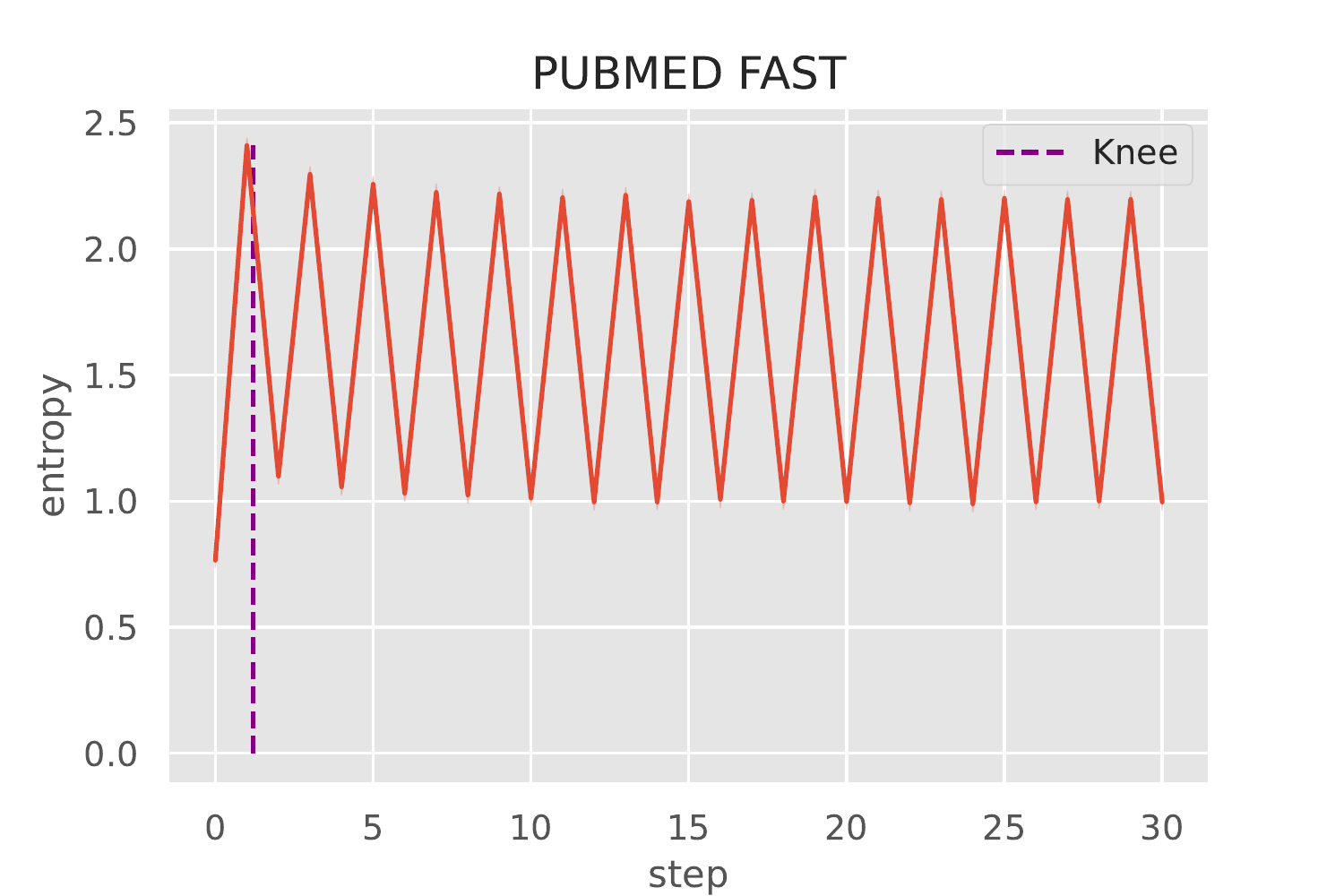}
\endminipage \\
\minipage{0.3\textwidth}
\includegraphics[width=\linewidth]{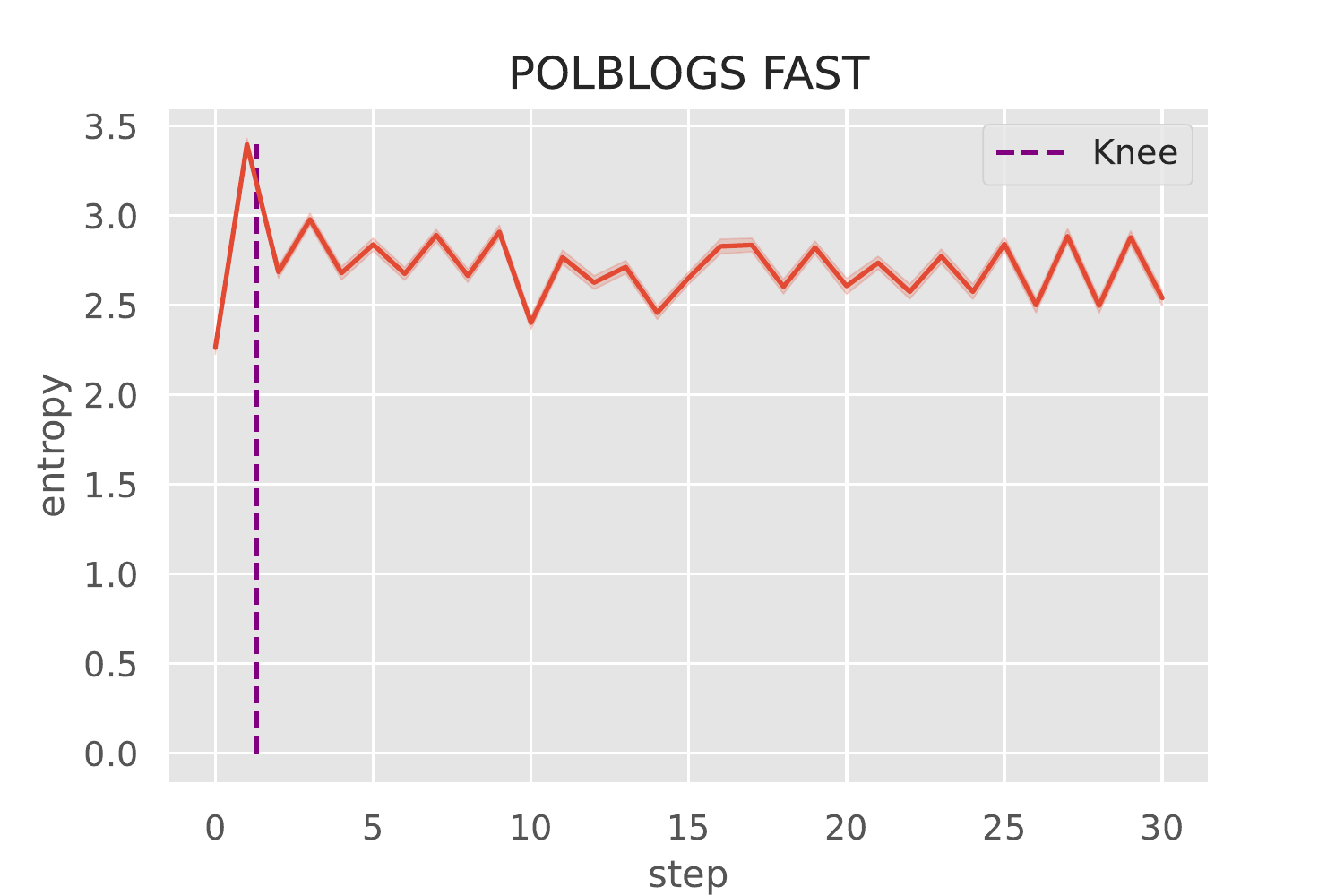}
\endminipage\hfill
\minipage{0.3\textwidth}
\includegraphics[width=\linewidth]{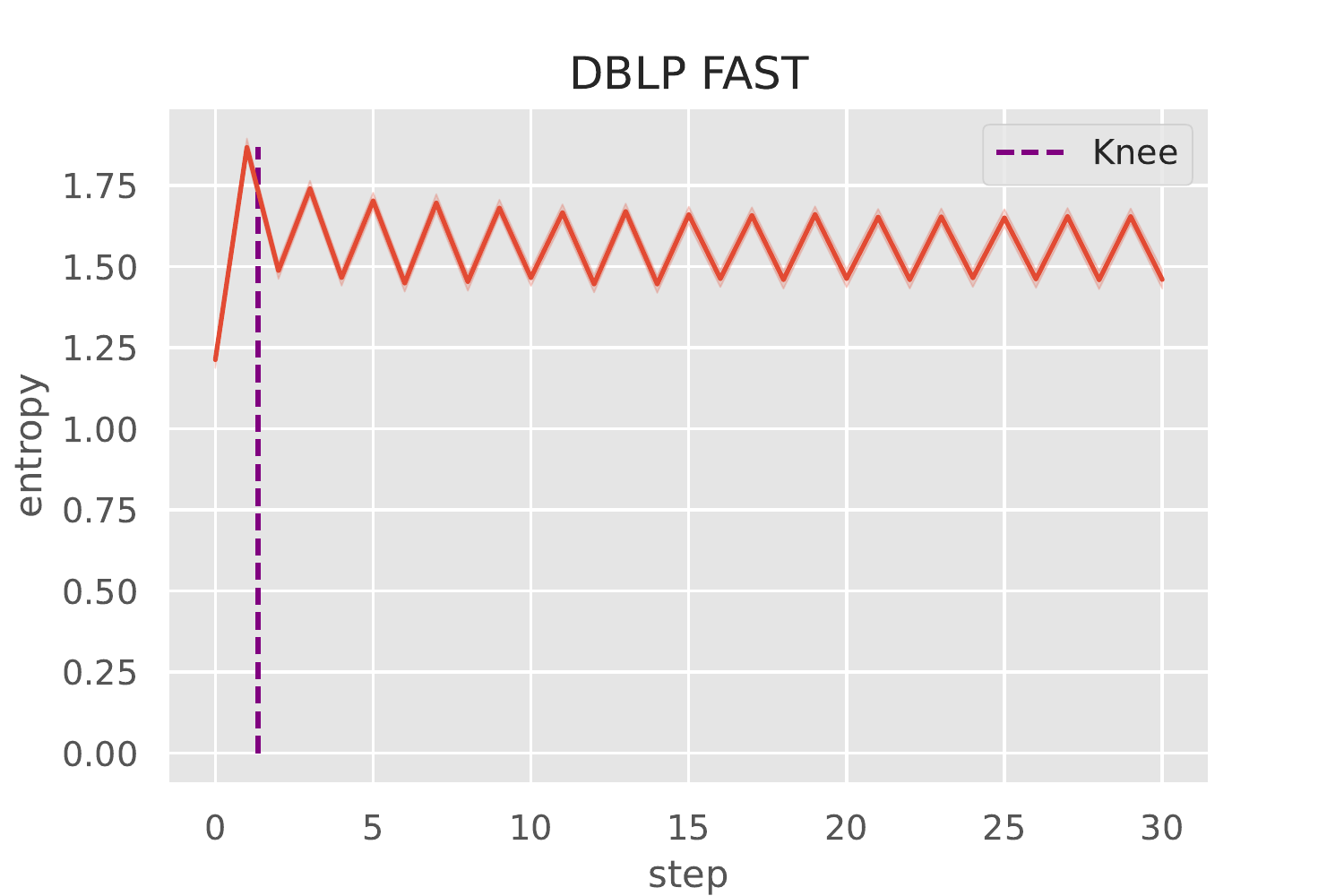}
\endminipage\hfill
\minipage{0.3\textwidth}%
\includegraphics[width=\linewidth]{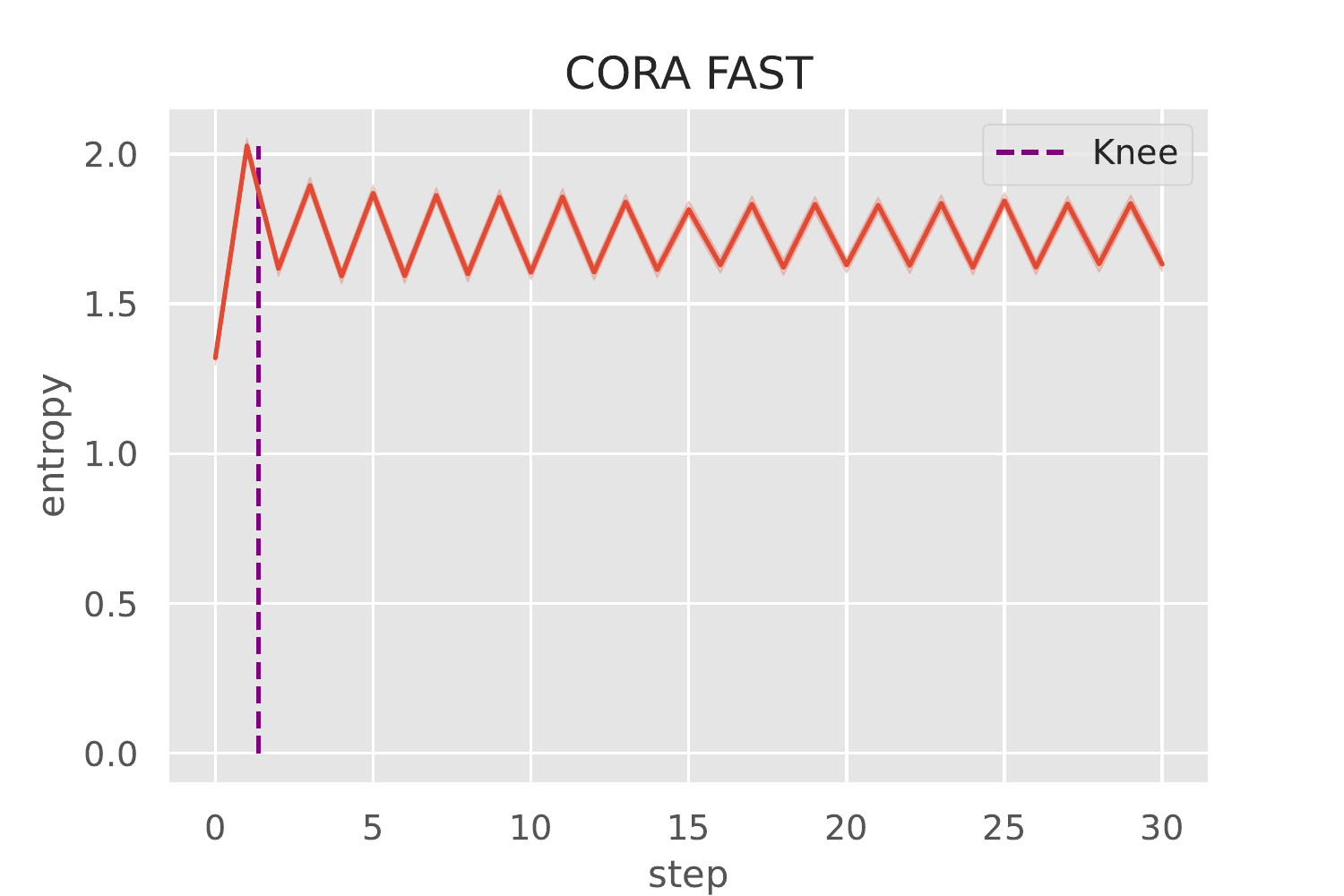}
\endminipage
\caption{Entropy analysis for the \fast models}
\label{fig:entropy_analysis_fast}
\end{figure*}

\begin{figure*}[!htb]
\minipage{0.3\textwidth}
\includegraphics[width=\linewidth]{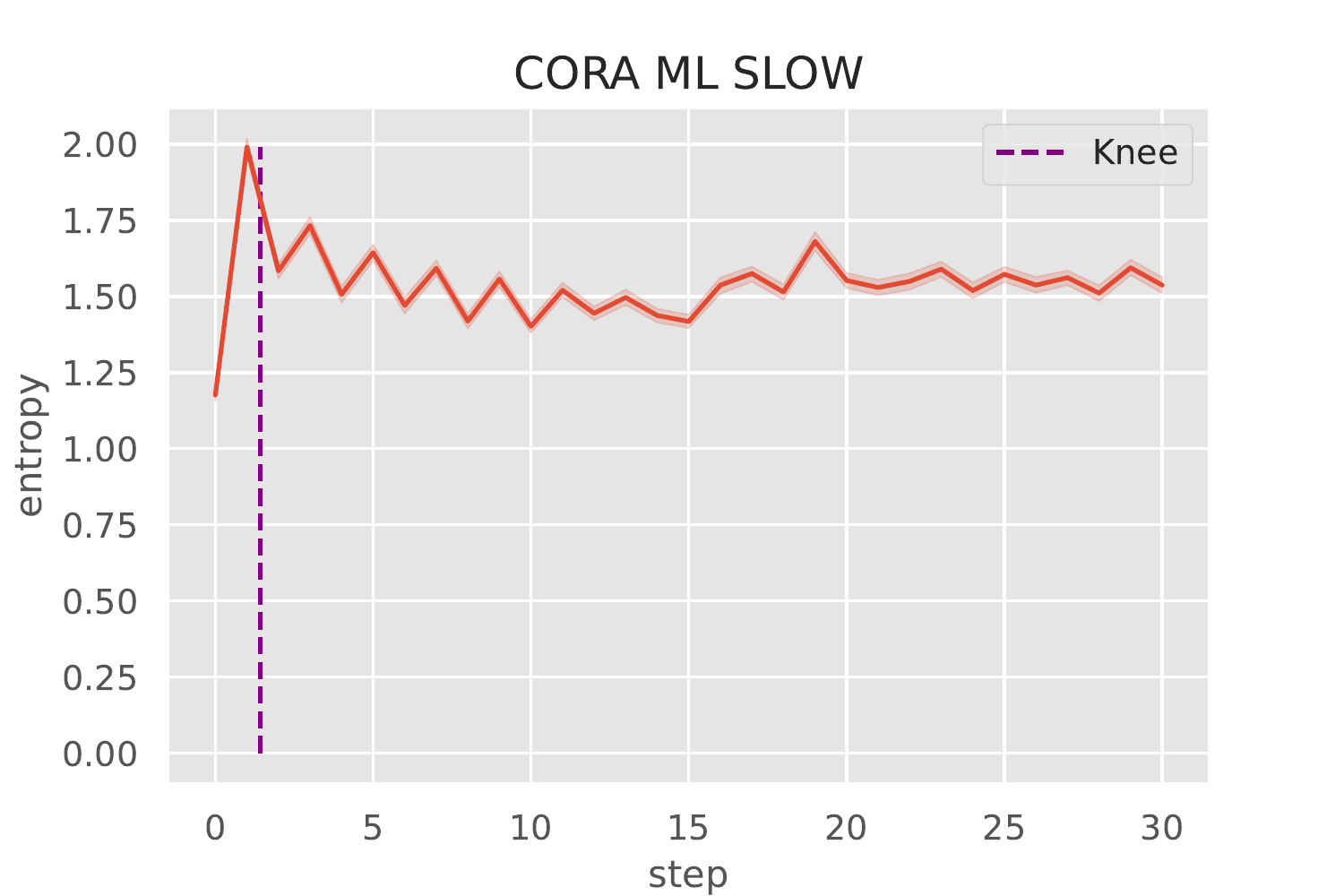}
\endminipage\hfill
\minipage{0.3\textwidth}
\includegraphics[width=\linewidth]{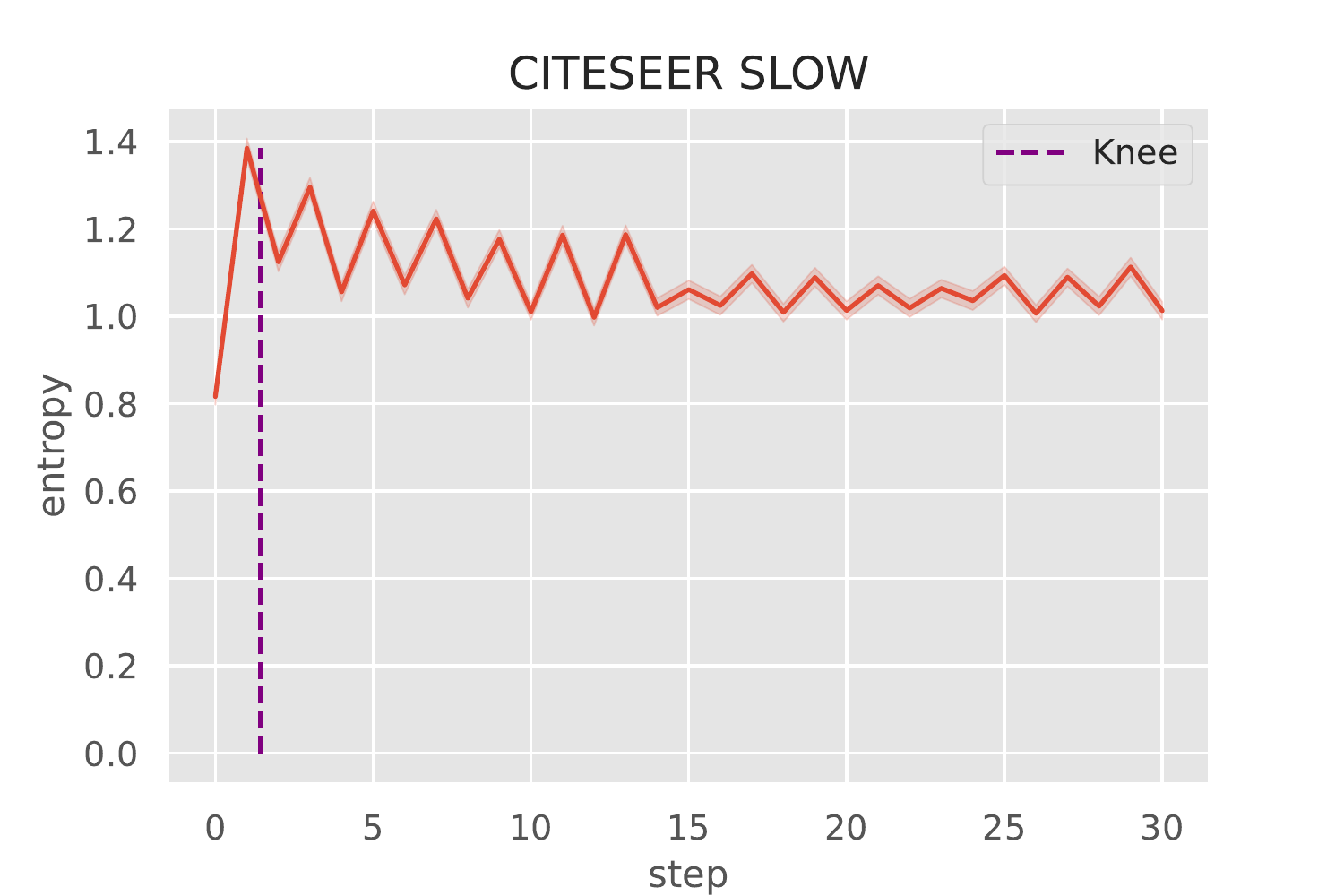}
\endminipage\hfill
\minipage{0.3\textwidth}
\includegraphics[width=\linewidth]{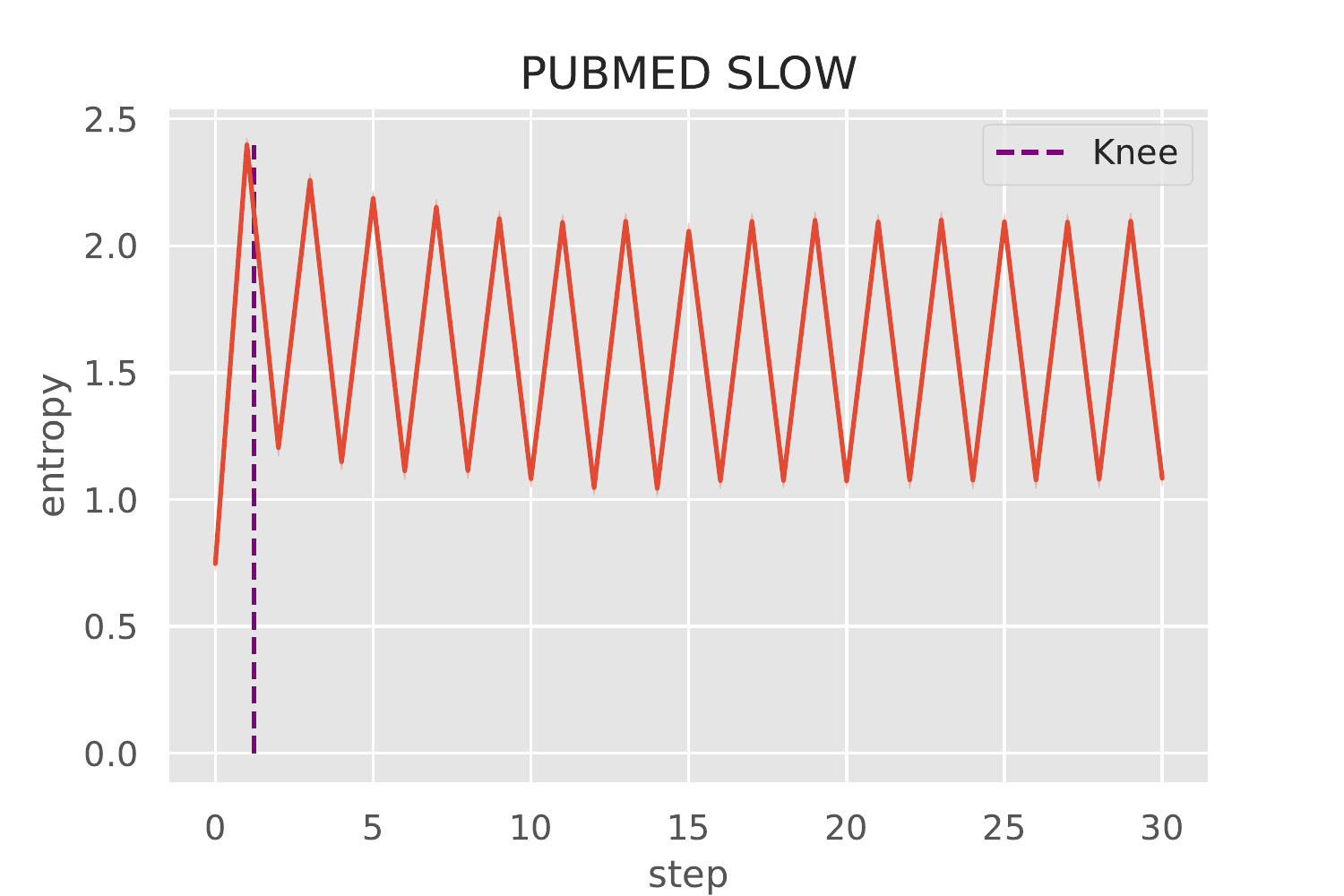}
\endminipage \\
\minipage{0.3\textwidth}
\includegraphics[width=\linewidth]{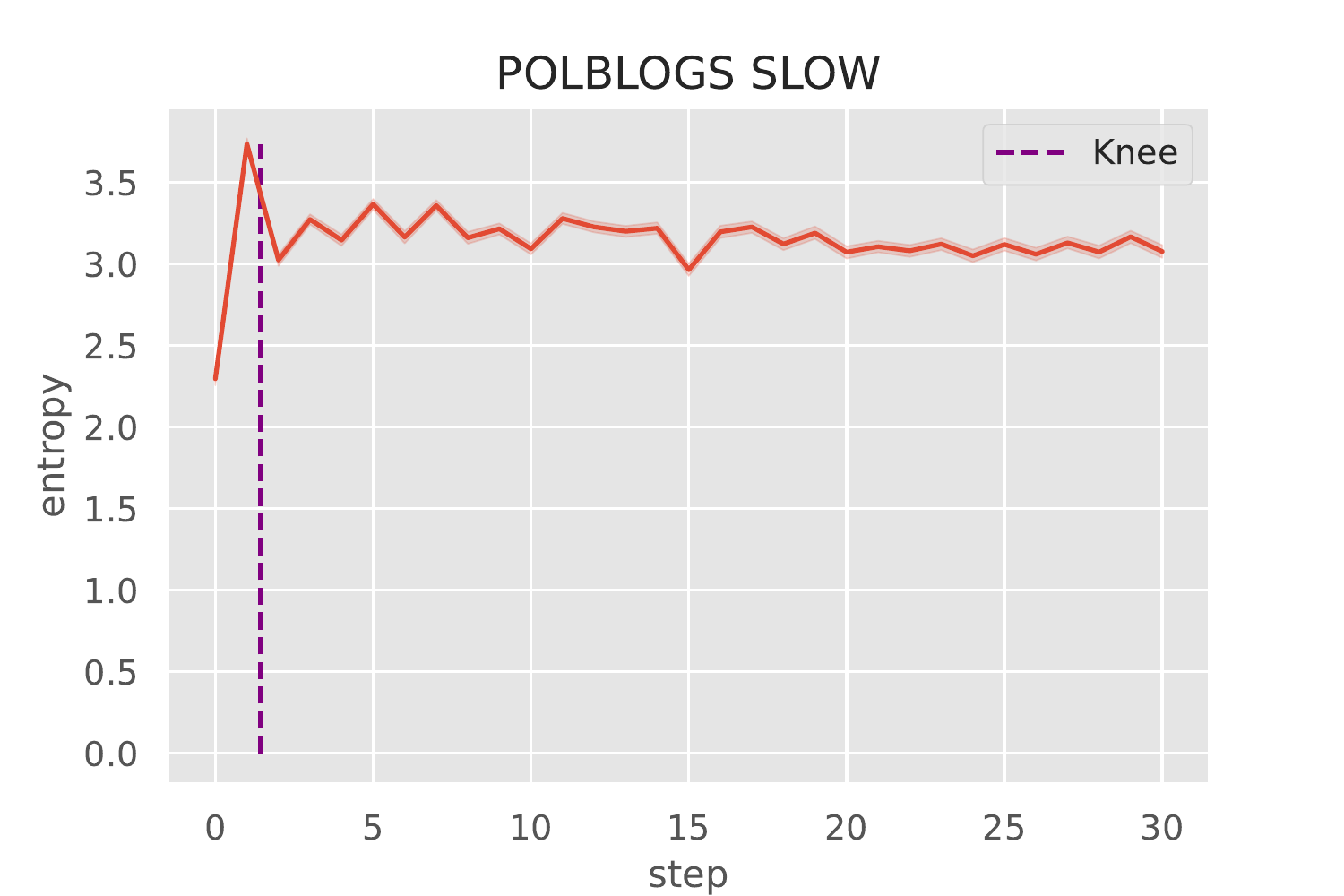}
\endminipage\hfill
\minipage{0.3\textwidth}
\includegraphics[width=\linewidth]{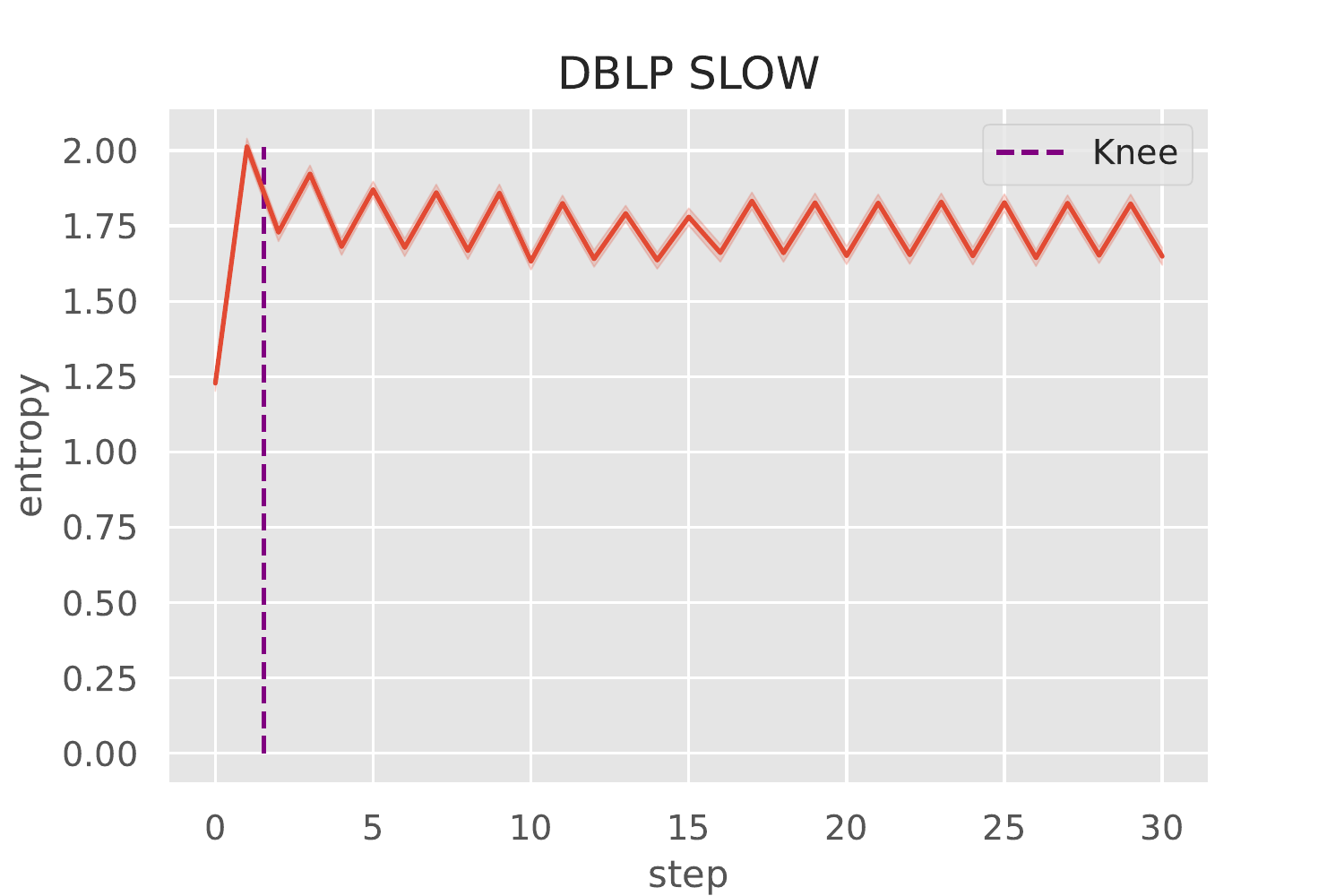}
\endminipage\hfill
\minipage{0.3\textwidth}%
\includegraphics[width=\linewidth]{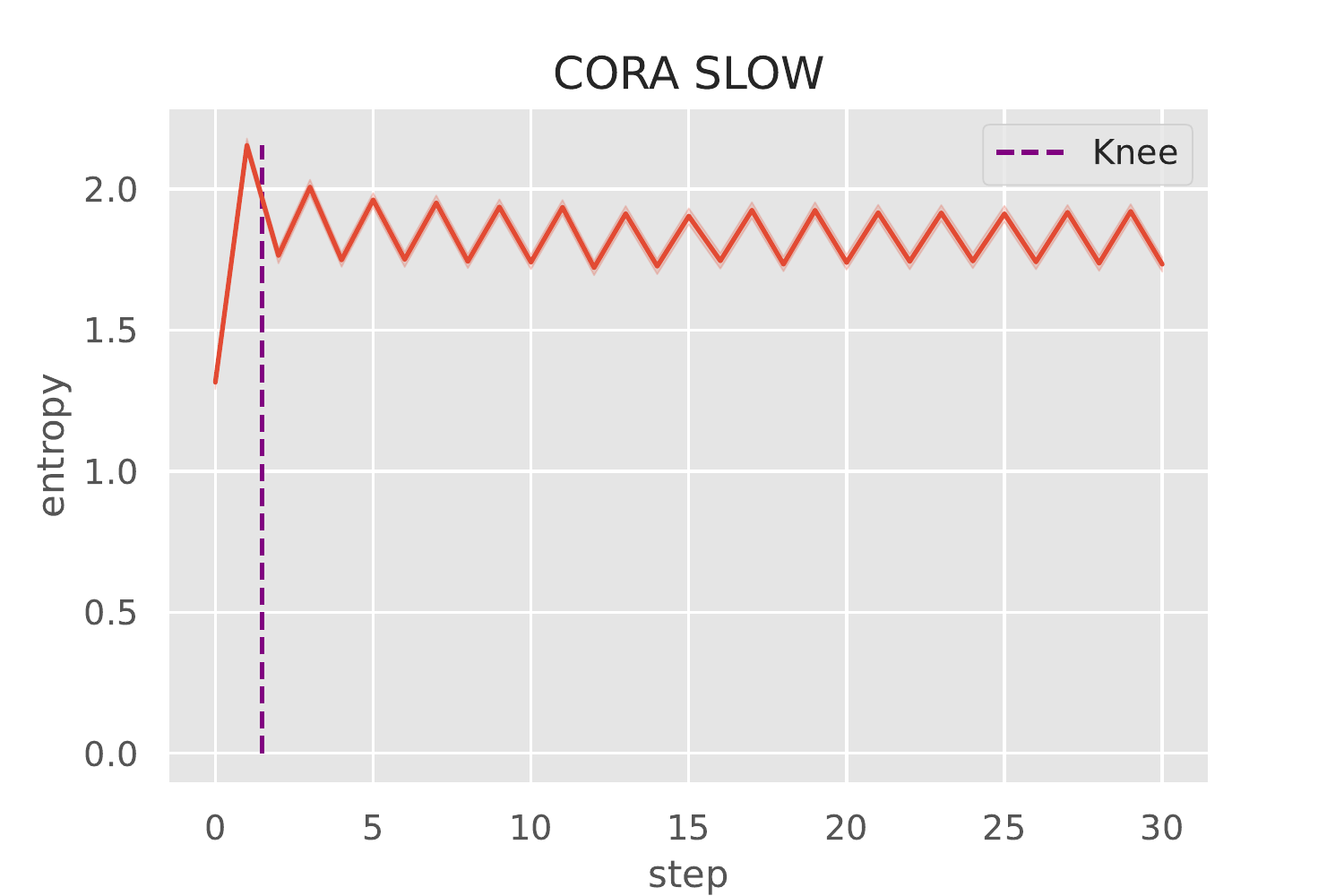}
\endminipage
\caption{Entropy analysis for the \slow models}
\label{fig:entropy_analysis_slow}
\end{figure*}

\aow{Specifically, we generate a walk of length 32, and for each step $i$, we calculate the entropy of the distribution $\pcond{\node_i}{\node_1, \node_2, \ldots, \node_{i-1}}$. The average entropy at each step is calculated, and the knee~\citep{satopaa2011finding} of the entropy plot is used as the switching point.
The results are shown in Figures~\ref{fig:entropy_analysis_fast} and \ref{fig:entropy_analysis_slow}.
As the Figures show, the knee point is detected early on for all the cases~(within ~4 steps), and then fluctuates around a mean value.
}

\end{document}